%% file: 00main.tex
\documentclass[10pt,twocolumn,letterpaper]{article}

\usepackage[pagenumbers]{cvpr} 
\makeatletter
\@namedef{ver@everyshi.sty}{}
\makeatother
\usepackage{graphicx}
\usepackage{amsmath}
\usepackage{amssymb}
\usepackage{booktabs}
\usepackage[utf8]{inputenc} 
\usepackage[T1]{fontenc}    
\usepackage{url}            
\usepackage{booktabs}       
\usepackage{amsfonts}       
\usepackage{nicefrac}       
\usepackage{microtype}      
\usepackage[dvipsnames]{xcolor}         
\usepackage{svg}

\usepackage{makecell}
\usepackage{multirow}
\usepackage{wrapfig}
\usepackage{xr}
\usepackage{adjustbox}
\usepackage{pifont}
\usepackage{enumitem}
\usepackage{subcaption}

\input{math_primitives}

\newcommand{\crelu}{$\mathbb{C}$ReLU}
\newcommand{\lab}{$L^*a^*b^*$}

\makeatletter
\newcommand{\printfnsymbol}[1]{%
        \textsuperscript{\@fnsymbol{#1}}%
}
\makeatother

\newcommand{\tb}[3]{\setlength{\tabcolsep}{#2mm}\begin{tabular}{#1}#3\end{tabular}}
\newcommand{\ol}[3]{\begin{#1}[leftmargin=*]\setlength{\itemsep}{#2mm}#3\end{#1}}

\usepackage{hyperref}
\hypersetup{pagebackref,breaklinks,colorlinks}

\usepackage[capitalize]{cleveref}
\crefname{section}{Sec.}{Secs.}
\Crefname{section}{Section}{Sections}
\Crefname{table}{Table}{Tables}
\crefname{table}{Tab.}{Tabs.}

\def\cvprPaperID{*} 
\def\confName{CVPR}
\def\confYear{2022}

\begin{document}

\title{Co-domain Symmetry for Complex-Valued Deep Learning}

\author{
\tb{ccc}{10}{
Utkarsh Singhal& 
Yifei Xing&
Stella X. Yu\\
}\\
UC Berkeley / ICSI\\
{\tt\small \{s.utkarsh, xingyifei2016, stellayu\}@berkeley.edu}
}

\maketitle

\input{figure_defn}

\input{table_defn}

\input{0abs}
\input{1intro}

\input{2related}
\input{3model}
\input{4experiments}

{\small
\bibliographystyle{unsrt}
\bibliography{references}
}

\input{6supp}
\end{document}

%% file: math_primitives.tex
\usepackage{amsthm,amsmath,mathtools}
\usepackage{enumitem}

\usepackage{color}
\usepackage{tabularx}
\usepackage{tabu}
\usepackage{multirow}
\usepackage{colortbl}
\usepackage{amssymb,booktabs}

\DeclareMathOperator{\arcdist}{arc}

\def\ang{\measuredangle}

%% file: figure_defn.tex
\def\figCoDomain#1#2{
\begin{figure}[#1]
    \centering
    \includegraphics[width=#2\textwidth, clip]{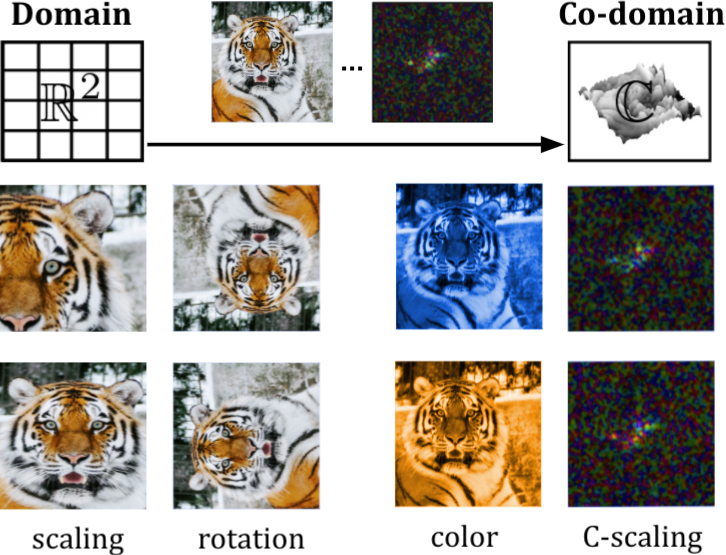}
    \caption{ An image is a function from the domain $\mathbb{R}^2$ to the co-domain $\mathbb{C}^N$. Image transformations like rotation and translation act on the domain, mapping points in $\mathbb{R}^2$ to other points, while leaving the underlying function values intact. Previous works like \cite{lieconv,cohen2016group} aim to produce architectures invariant to domain transformations. Co-domain transformations like color distortion or complex-valued scaling, on the other hand, act on the function values only.}
    \label{fig:co_domain}
\end{figure}
}

\def\figInvariance#1{
\begin{figure*}[#1]
    \centering
    \setlength{\tabcolsep}{0pt}
    \begin{subfigure}{0.9\textwidth}  \centering \includegraphics[width=0.9\linewidth,trim=0 220 0 0, clip]{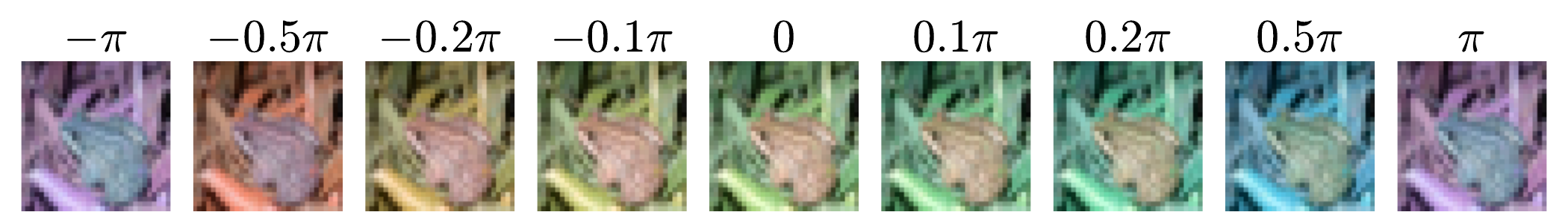} \caption{Color variations due to complex-scaling in our proposed LAB encoding } \label{fig:colorvar} \end{subfigure}
    \begin{tabular}{ccc}
    \begin{subfigure}{0.33\textwidth} \centering \includegraphics[height=3.5cm]{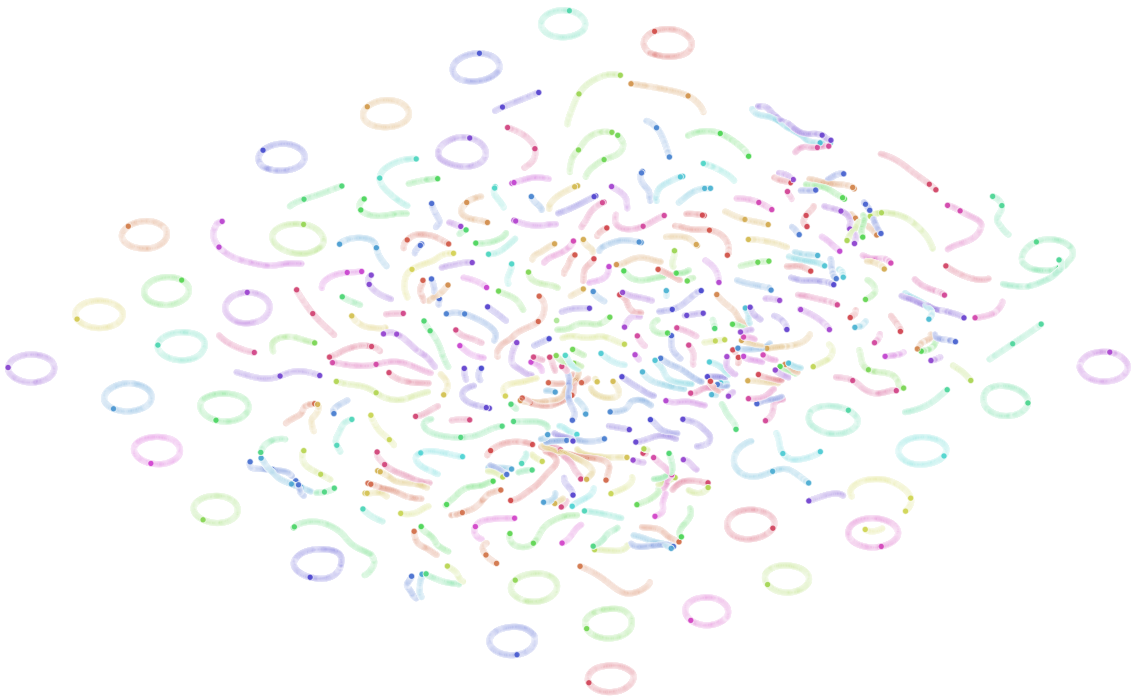} \caption{t-SNE Embedding for DCN \cite{trabelsi2017deep}} \label{fig:dcn_tsne} \end{subfigure} &   
    \begin{subfigure}{0.33\textwidth} \centering \includegraphics[height=3.5cm]{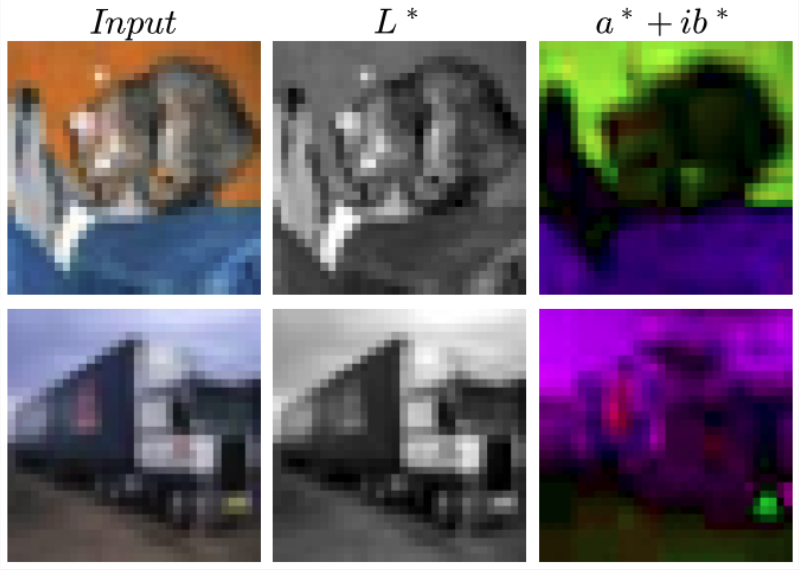}\caption{Visualization of our LAB encoding} \label{fig:labex} \end{subfigure}  & 
    \begin{subfigure}{0.33\textwidth} \centering \includegraphics[height=3.5cm]{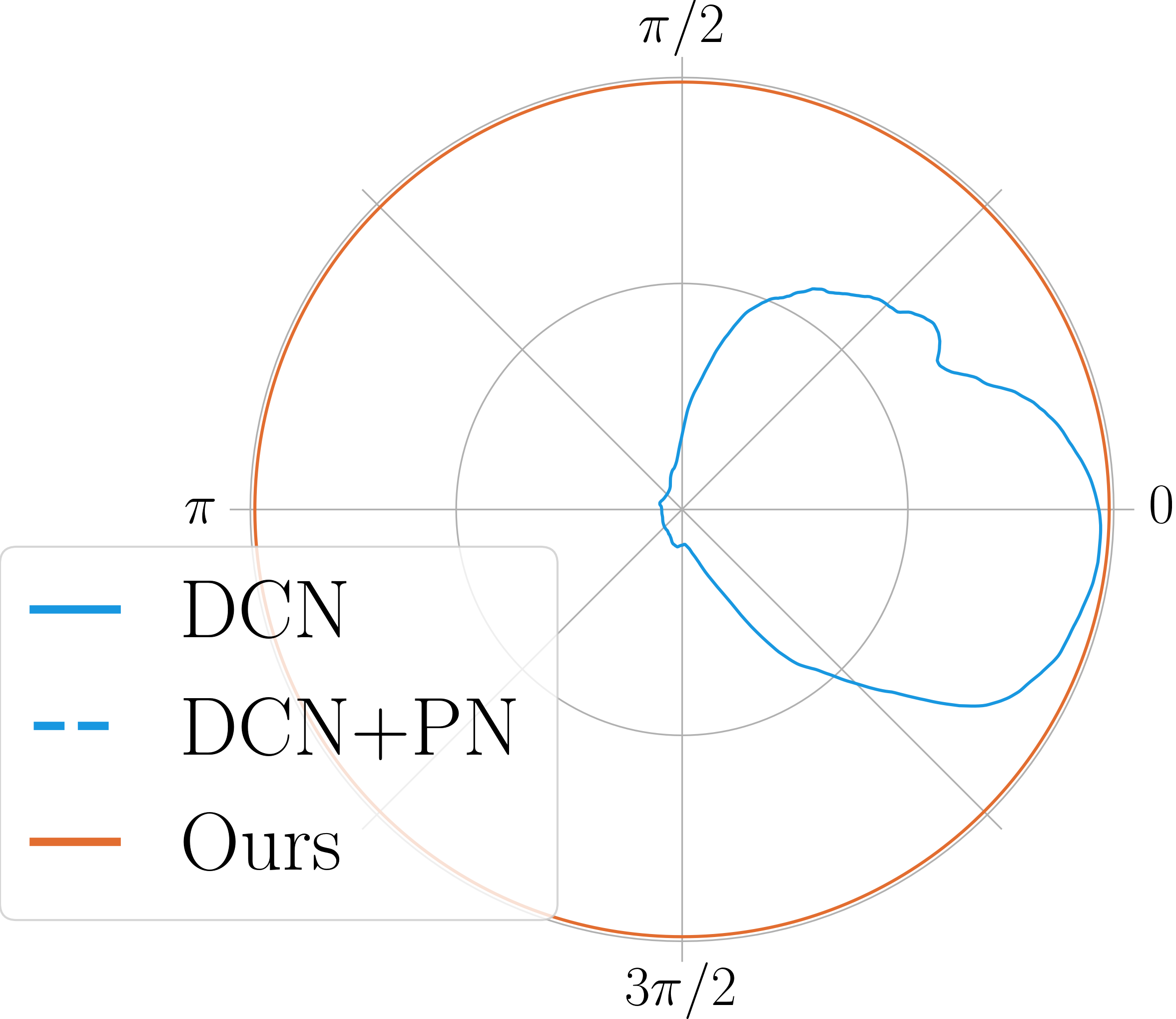}\caption{Model confidence for a single example} \label{fig:polar} \end{subfigure}\\
    \begin{subfigure}{0.33\textwidth} \centering \includegraphics[height=3.5cm]{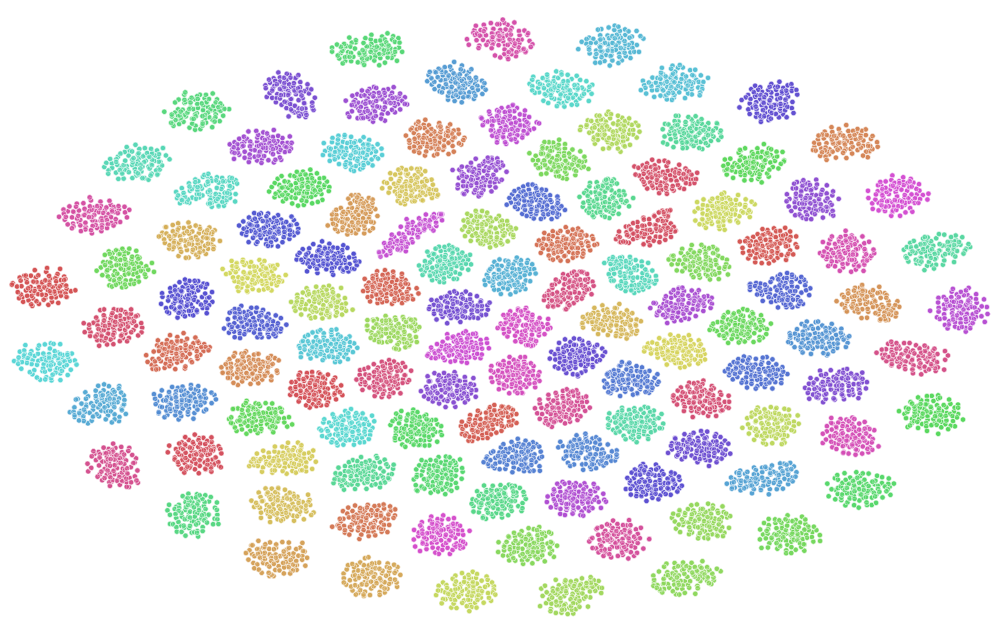}  \caption{t-SNE Embedding for our model} \label{fig:our_tsne} \end{subfigure} & 
    \begin{subfigure}{0.33\textwidth} \centering \includegraphics[height=3.5cm]{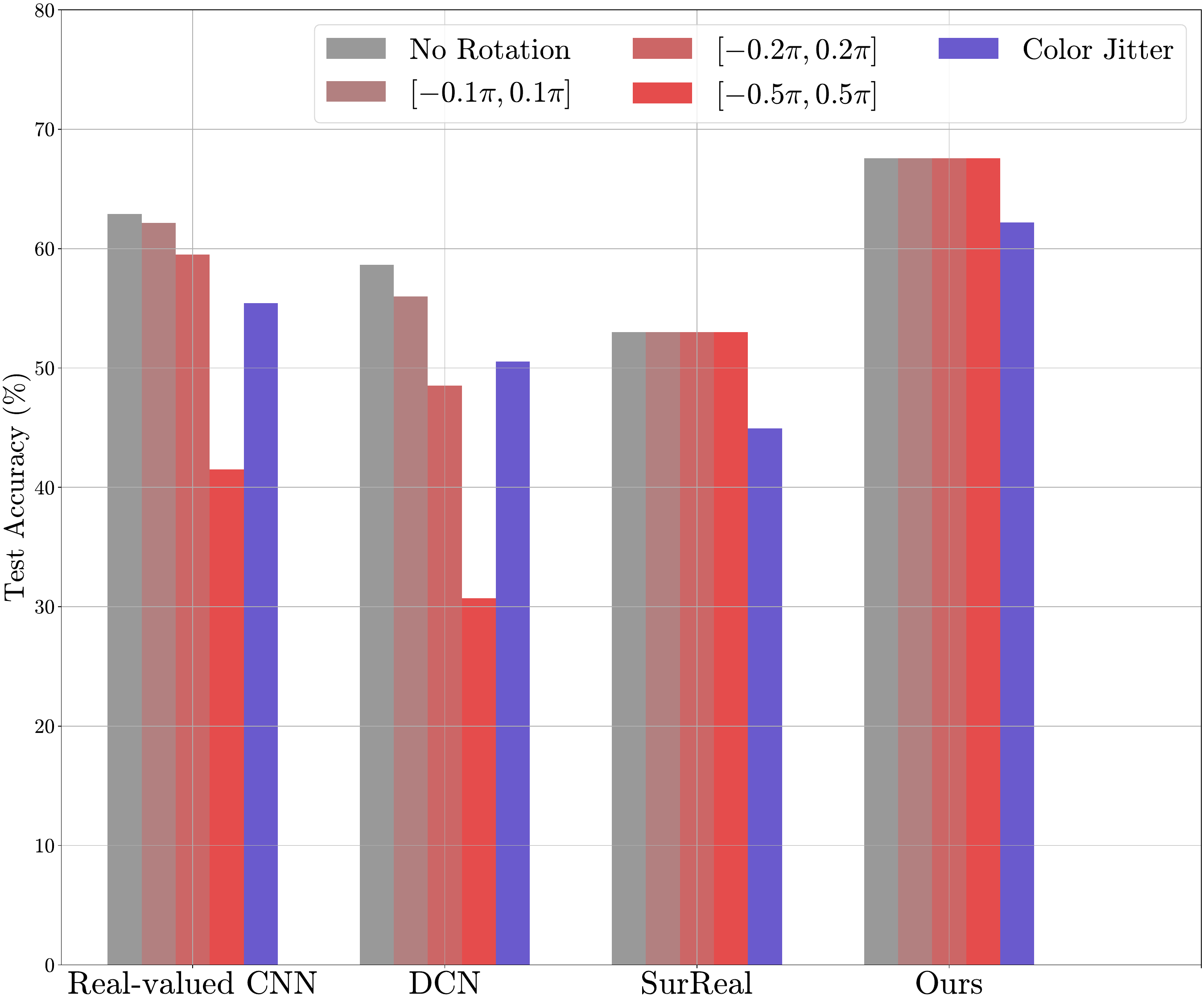}  \caption{Accuracy under color jitter} \label{fig:coljit} \end{subfigure} &
    \begin{subfigure}{0.33\textwidth} \centering \includegraphics[height=3.5cm]{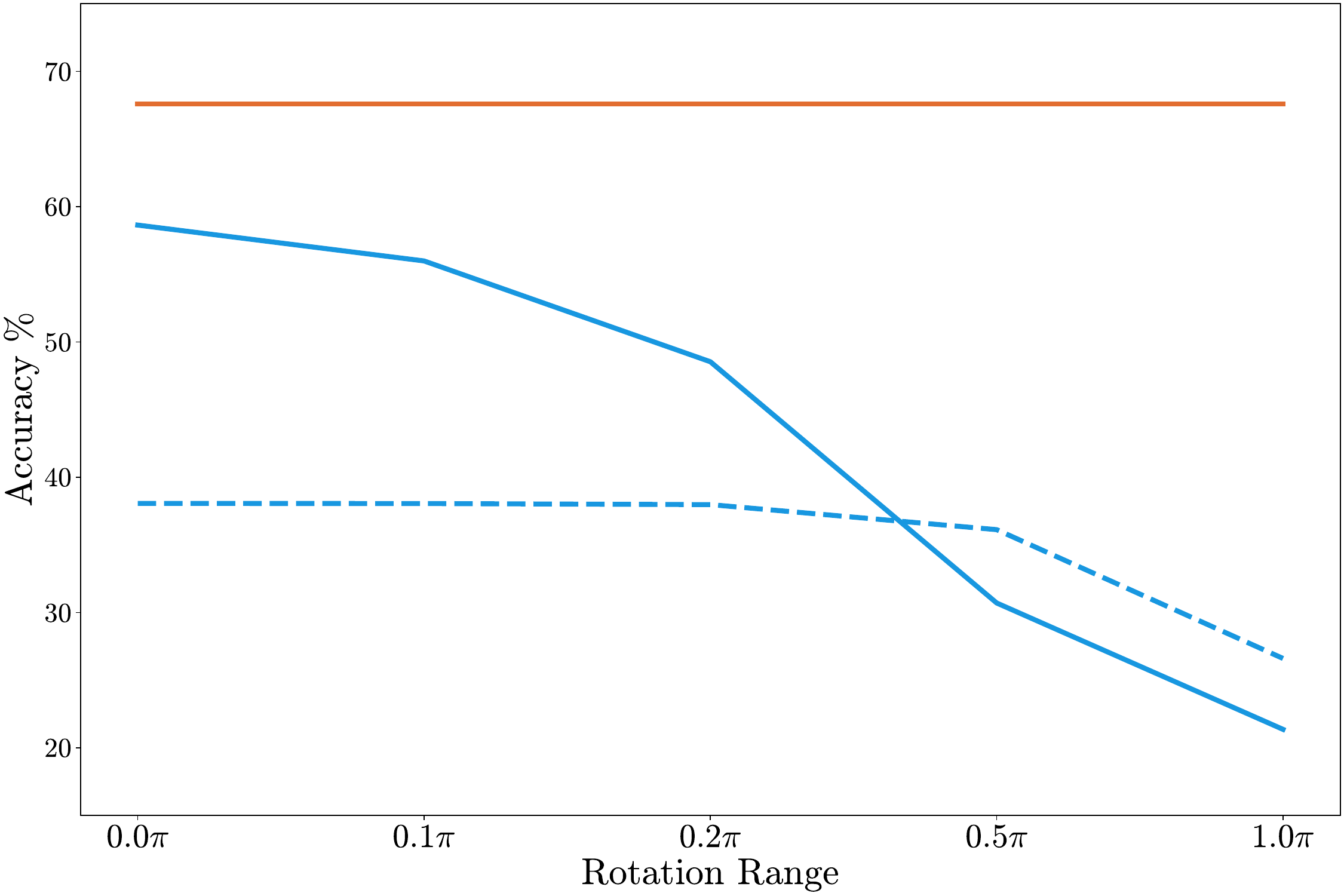}  \caption{Comparison against phase normalization} \label{fig:PN} \end{subfigure}
    \end{tabular}
    \caption{Our method learns invariant features with respect to complex-scaling of the input. All examples are from CIFAR 10 with our LAB encoding, undergoing multiplication by a unit complex number. \textbf{(b, e)} tSNE embedding trajectories from DCN \cite{trabelsi2017deep} and our model. Each color represents a different example. Embeddings form tight clusters for our model, and irregular overlapping curves for DCN. \textbf{(c)} Visualization of our complex-valued embedding of LAB information. The $L^*$ channel is visualized as a grayscale image, and the complex-valued $a^* + \textit{i} b^*$ visualized as a color image. \textbf{(d)} Model confidence of the correct class for a single example. Higher confidence means larger radius. DCN predictions are highly variable, while our model is robust to complex-scaling and thus constant. \textbf{(f)} Accuracy under complex-scaling and color jitter. Red bars represent complex-rotations sampled from different rotation ranges. Blue bars represent color jitter (as used in \cite{mean_teach}). Our method maintains high accuracy across complex-rotations and color jitter, whereas DCN and Real-valued CNN fail. SurReal \cite{chakraborty2019surreal} is robust, but has low overall accuracy. Our method combines high accuracy with robustness. \textbf{(g)} Average accuracy under different rotation ranges, comparing DCN with phase normalization (dotted blue line) and without phase normalization (solid blue line) against our method. The color encoding has a complicated phase distribution, and phase normalization fails to estimate the amount of rotation, resulting in poor accuracy. Our method is thus more suitable for complicated phase distributions.}
    \label{fig:invariance}
\end{figure*}
}

\def\figGeneralization#1{
\begin{figure*}[#1]
    \centering
    \setlength{\tabcolsep}{1pt}
    \begin{tabular}{@{}ccc@{}}
    \begin{subfigure}[b]{0.33\textwidth} \centering \label{fig:scaling} \includegraphics[height=4.3cm]{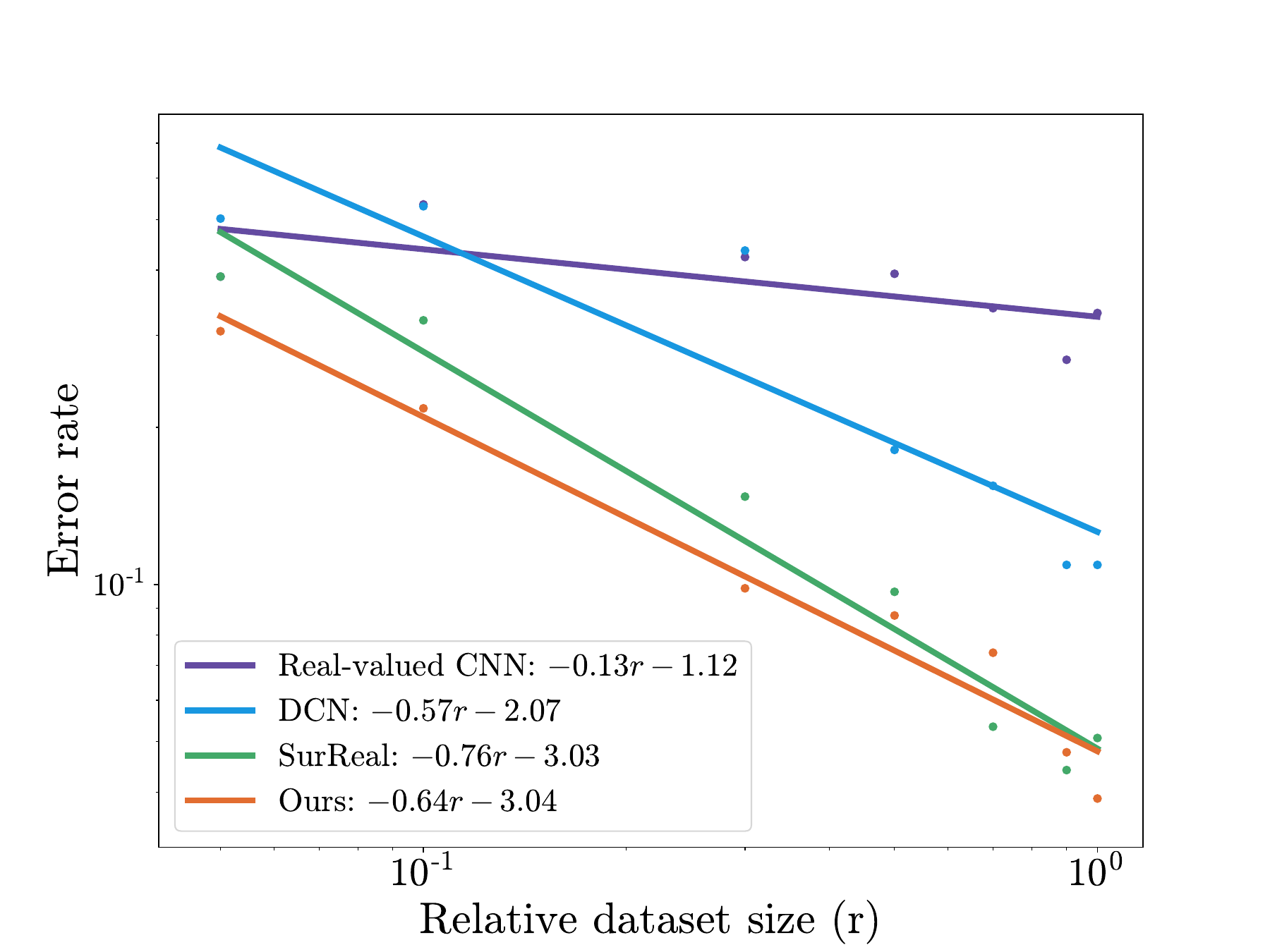}    \caption{Scaling Laws for MSTAR \cite{kaplan2020scaling}} \end{subfigure} &   
    \begin{subfigure}[b]{0.33\textwidth} \centering \includegraphics[height=3.9cm]{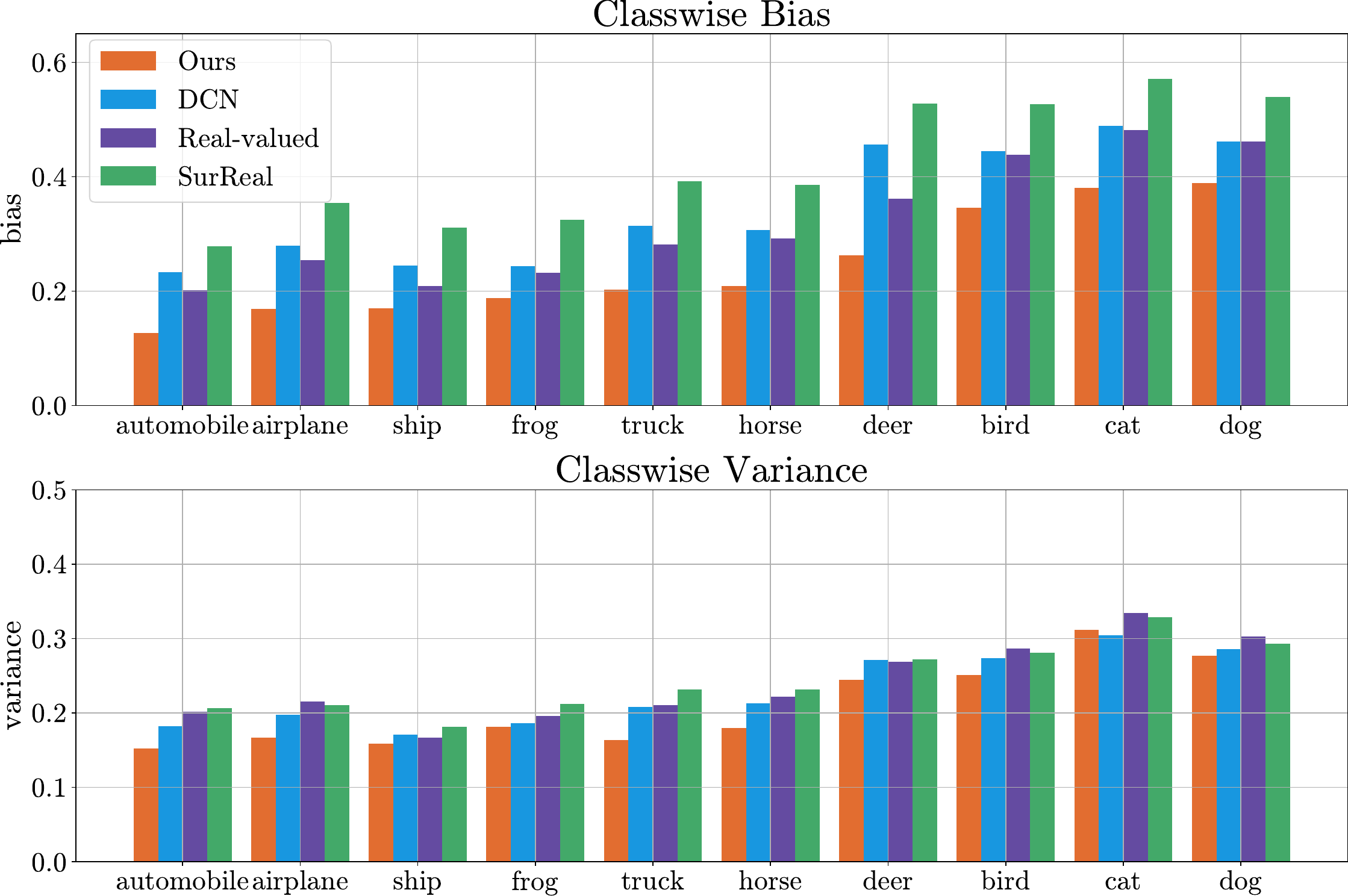}  \label{fig:biasvar}  \caption{Class-wise Bias and Variance for CIFAR10 \cite{wang2021longtailed}} \end{subfigure} &  
    \begin{subfigure}[b]{0.33\textwidth} \centering \includegraphics[height=4.3cm]{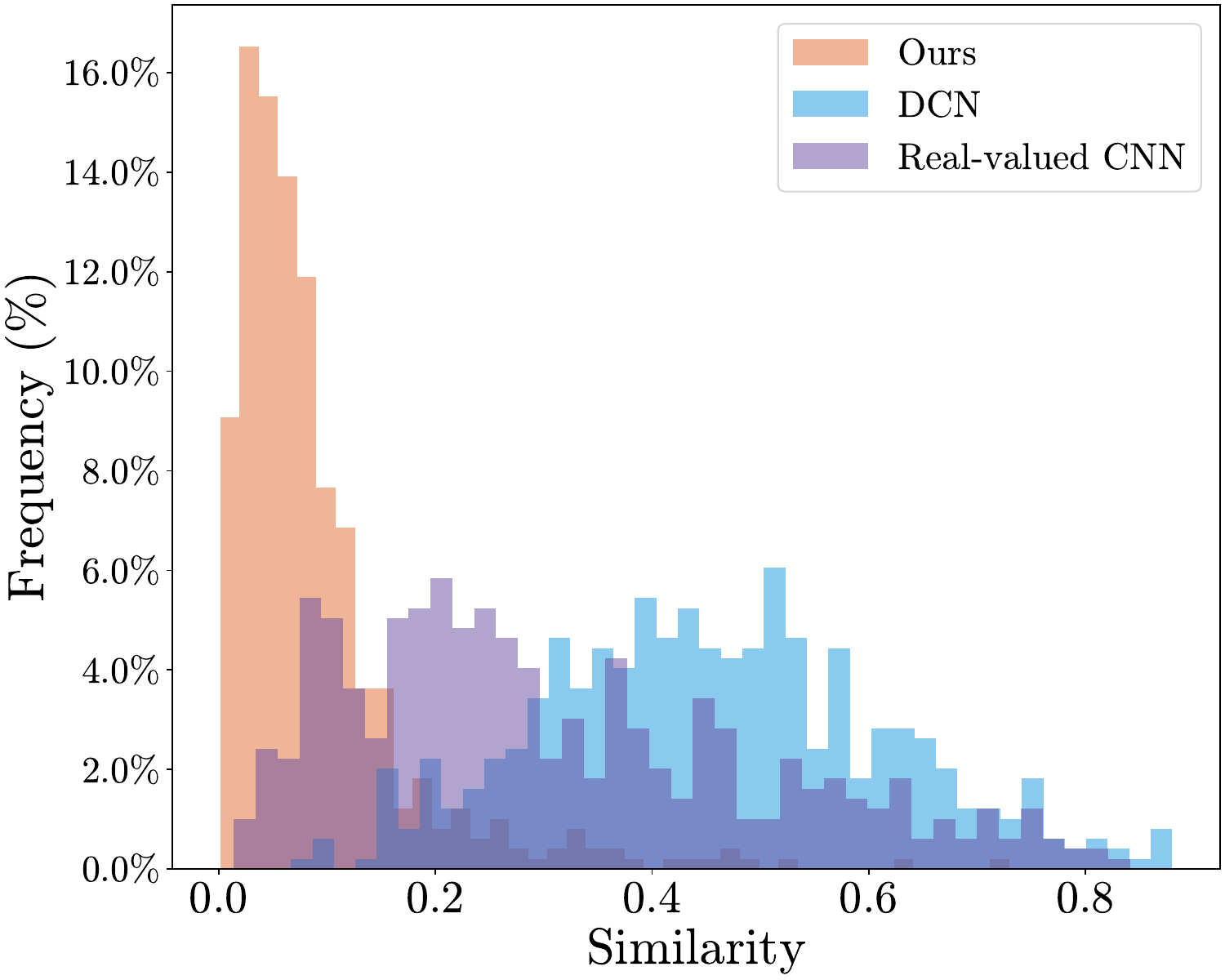} \label{fig:filtsim}   \caption{Filter similarity histogram for \textit{conv2}} \end{subfigure}  \\

    \end{tabular}
    \vspace*{-3mm}
    \caption{Our model generalizes across various dataset sizes, has lower bias and variance, and avoids learning redundant filters. \textbf{(a)}: We produce trend curves (similar to \cite{kaplan2020scaling}) for the MSTAR accuracy table (Table \ref{table:mstar_result}). Our method has the lowest test error for measured dataset sizes, a trend that is predicted to scale to even smaller sizes. \textbf{(b)}: We followed \cite{wang2021longtailed} for CIFAR10 models with LAB encoding. Classes are ordered in ascending order of bias for our model. Our model consistently shows the lowest bias for each class, and the lowest variance for 9 out of 10 classes, indicating overall superior generalization ability. \textbf{(c)}: Filter similarity histogram from \textit{conv2} layer of each CIFARnet model, following \cite{wang2020tied}. Our distribution mean is closest to $0$, indicating our method achieves the least redundant filters.}
    \label{fig:generalization}
\end{figure*}
}

\def\figgtrelu#1{
\begin{figure}[#1]
    \centering
    \setlength{\tabcolsep}{0pt}
    \begin{tabular}{cccc}
    \begin{subfigure}[b]{2.1cm} \centering \includegraphics[height=2.1cm]{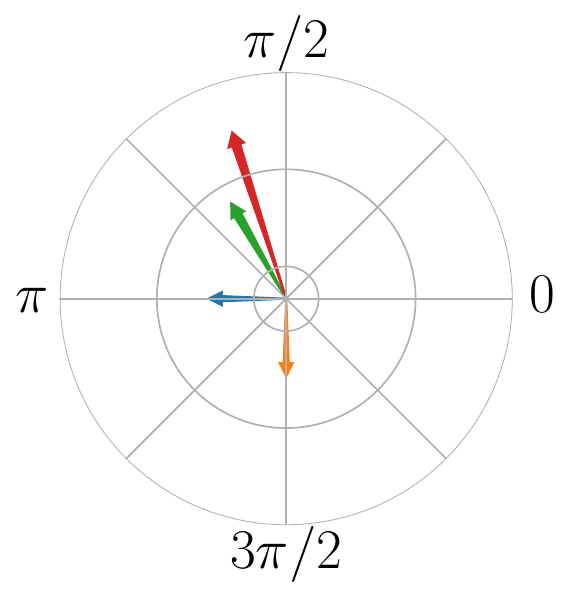} \caption{Input features} \end{subfigure} &
    \begin{subfigure}[b]{2.1cm} \centering \includegraphics[height=2.1cm]{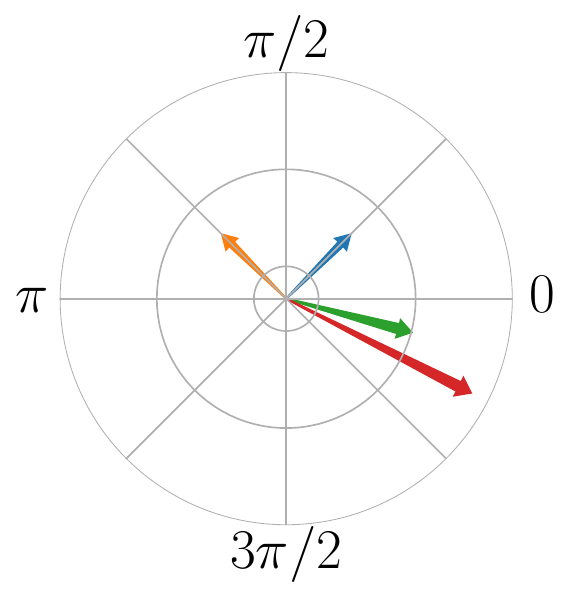} \caption{Scaling factor} \end{subfigure}&
    \begin{subfigure}[b]{2.1cm} \centering \includegraphics[height=2.1cm]{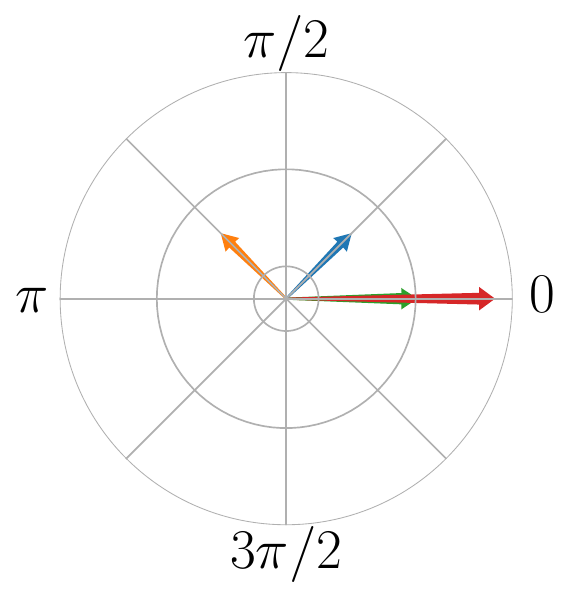} \caption{Thresholding} \end{subfigure}&
    \begin{subfigure}[b]{2.1cm} \centering \includegraphics[height=2.1cm]{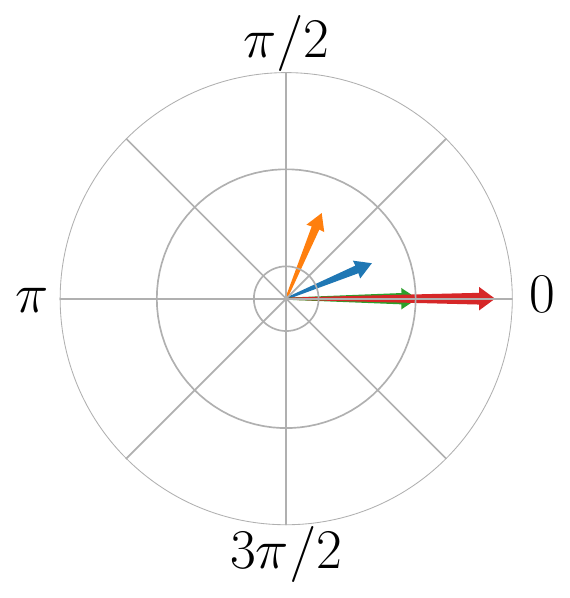}  \caption{Phase-scaling} \end{subfigure}
    \end{tabular}
    \caption{Our Generalized Tangent ReLU transforms the input in three stages: \textbf{(a)} given input complex vectors, it multiplies each channel with a learned scaling factor, \textbf{(b)} thresholds the input magnitude and phase with hyperparameter $r$, and \textbf{(c)} scales the phase to adapt the output distribution.}
    \label{fig:gtrelu}
\end{figure}
}

\def\figmodels#1{
\begin{figure*}[#1]
    \centering
    \setlength{\tabcolsep}{0pt}
\tb{@{}l|c@{}}{2}{
\bf\tb{c}{0}{Model\\Type I} & 
\tb{c}{0}{
\includegraphics[height=2.8cm]{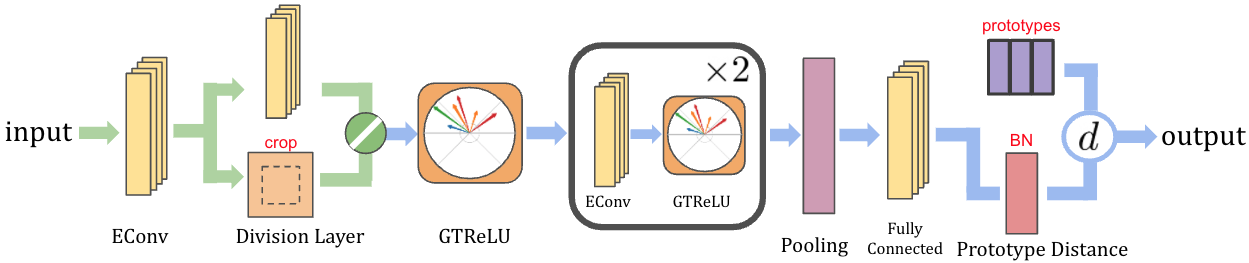}
}\\ \midrule
\bf\tb{c}{0}{Model\\Type E} & 
\tb{c}{0}{
\includegraphics[height=2.8cm]{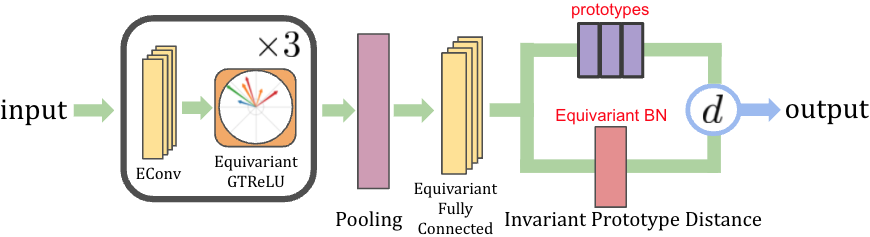}
}\\
}
    \caption{Our CIFARnet models demonstrate two methods of constructing complex-scale invariant models. Green arrows represent equivariant features, and blue arrows represent invariant features. \textbf{top:} \textbf{Type I} architecture uses a \textit{Division Layer} in early stages, producing complex-scale invariant features which can be used with any following layers. \textbf{bottom:} \textbf{Type E} uses equivariant layers throughout the network, retaining phase information until the final \textit{Invariant Prototype Distance Layer}. This class of models is more restrictive but can achieve higher accuracy (See Table \ref{table:conv_compare_all}) as a consequence of retaining more information.} 
    \label{fig:models}
\end{figure*}
}

\def\figeqnl#1{
\begin{figure}[#1]
    \centering
    \setlength{\tabcolsep}{0pt}
    \begin{tabular}{cccc}
    \includegraphics[width=\linewidth]{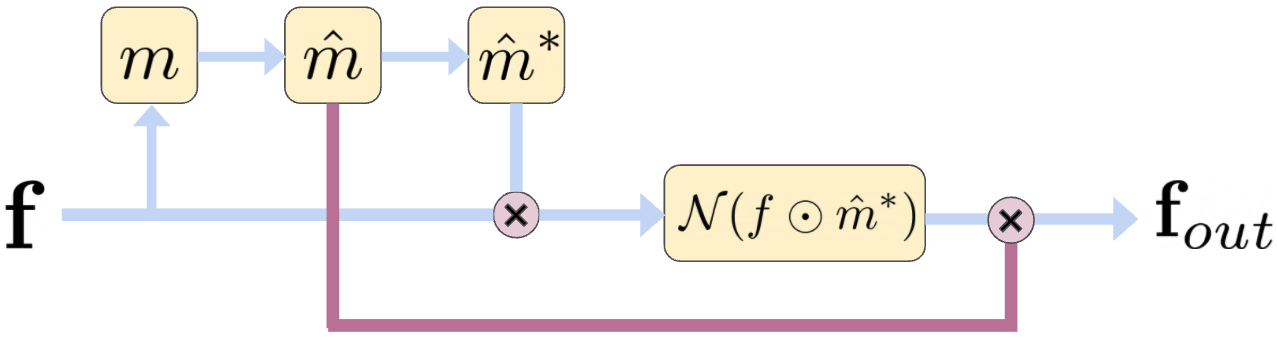} &
    \end{tabular}
    \caption{Our equivariant non-linearity, $\mathcal{E}\{\mathcal{N}\}$, works in four stages. We compute the channel mean $m$ of the input feature $f$ and normalize it to retain only phase information. This normalized mean vector $\hat{m}$ is equivariant to input phase. We multiply $f$ by $\hat{m}^*$ to cancel the input phase, resulting in a phase-invariant feature $f \odot \hat{m}^*$. We feed this feature to the non-linearity $\mathcal{N}$ and multiply by $\hat{m}$ to restore the previously cancelled input phase. The result is equivariant in phase and also equivariant in magnitude if $\mathcal{N}$ is.}
    \label{fig:enql}
\end{figure}
}

\def\figBigModels#1{
\begin{figure*}[#1]
    \centering
    \setlength{\tabcolsep}{0pt}
    \begin{tabular}{cc}
    \includegraphics[width= 0.3 \textwidth]{fig/mstar_scaling.pdf}  & \includegraphics[width= 0.3 \textwidth]{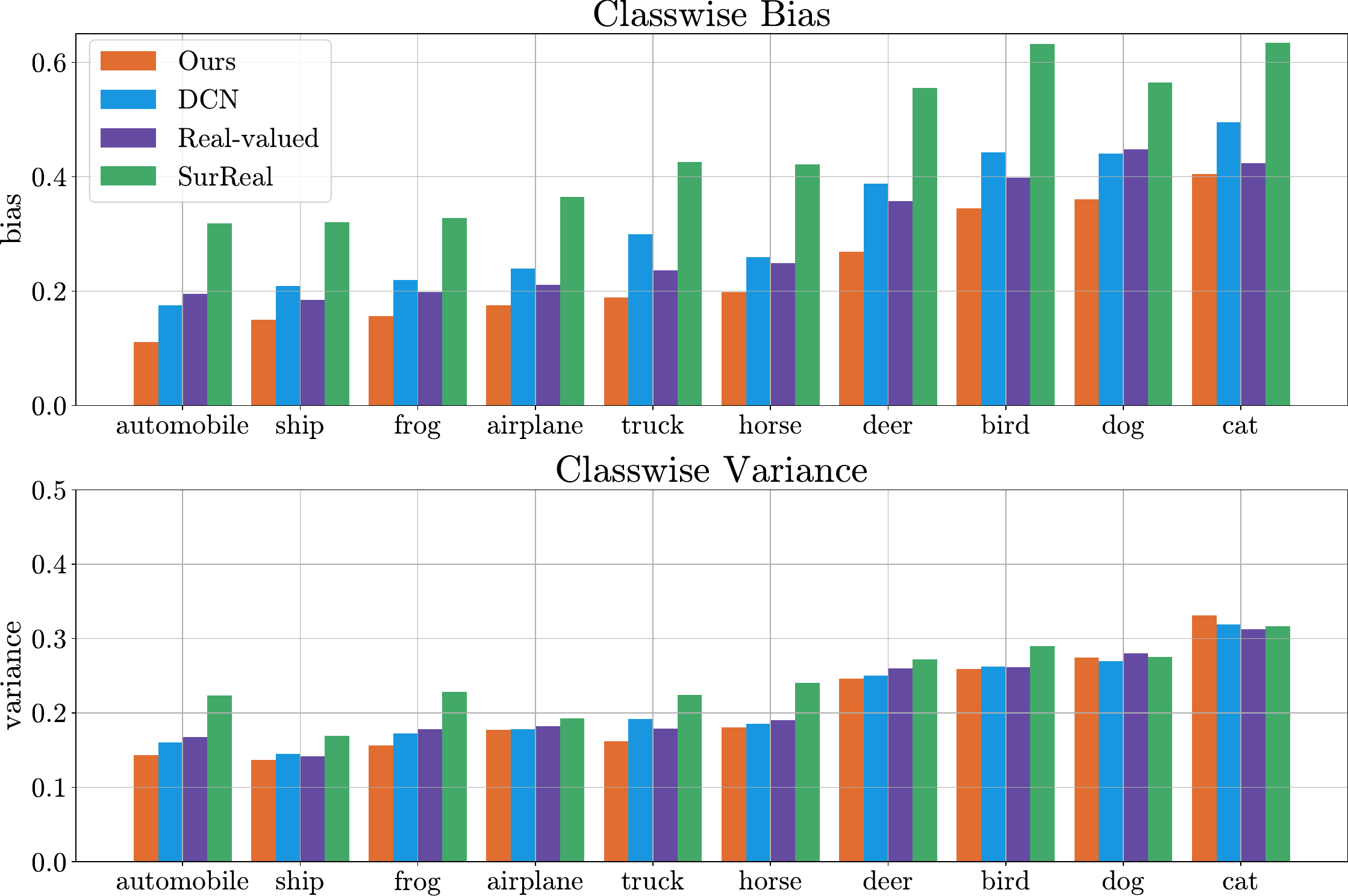} \\
     \includegraphics[width= 0.2 \textwidth]{fig/calibration.png} & \includegraphics[width= 0.4 \textwidth]{fig/filt_sim_2.png} \\
    \end{tabular}
    \caption{Our approach scales to larger models. In table (a) we compare the accuracy of our }
    \label{fig:generalization}
\end{figure*}
}

\def\figDivCinv#1{
\begin{figure}[#1]
    \centering
    \includegraphics[width=0.27\textwidth,clip]{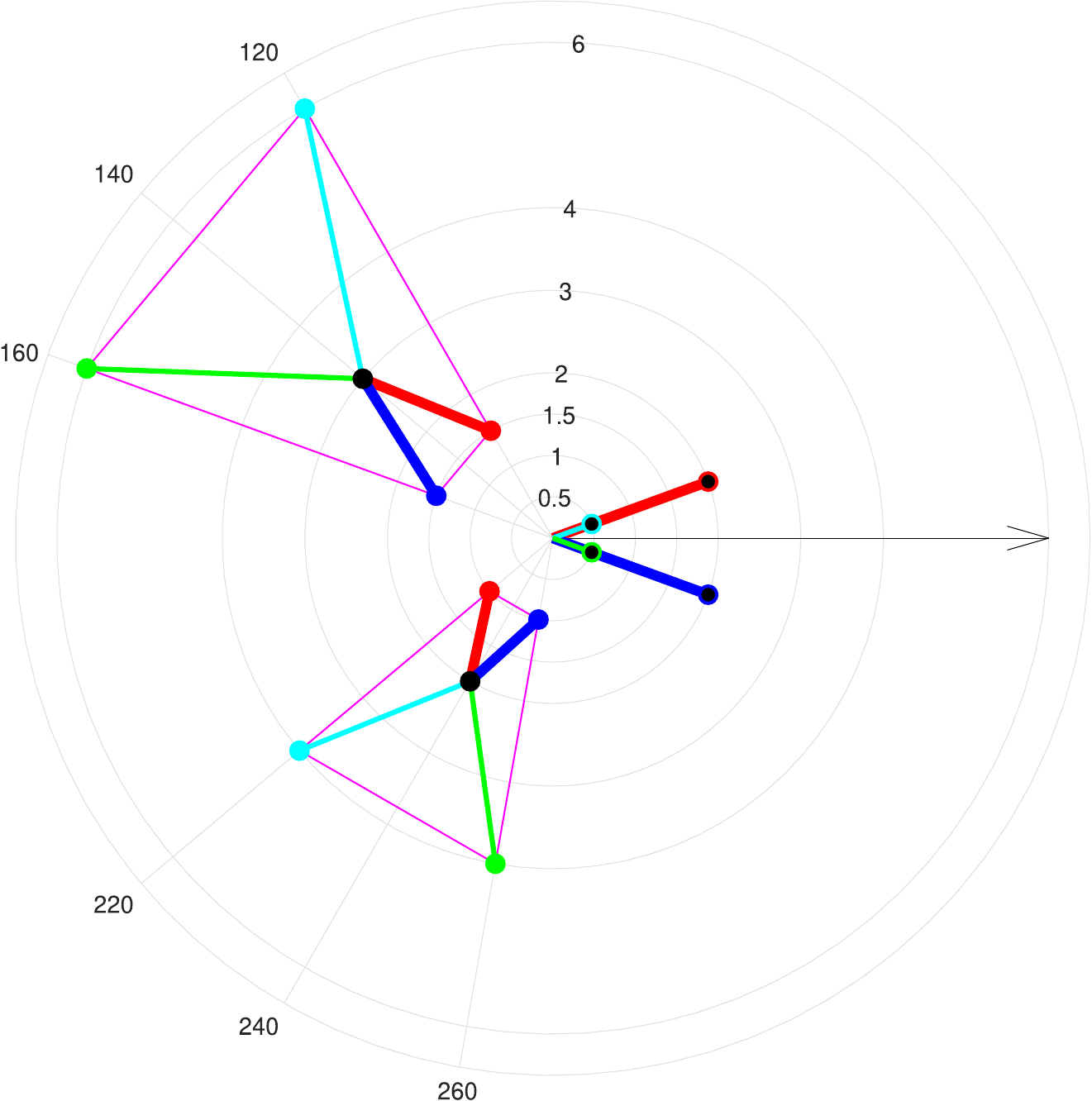}
    \caption{Our Division Layer is complex-scale invariant, and preserves more information than the Distance Transform \cite{chakraborty2019surreal}. Consider 4 complex numbers (colored dots) and the reference (black dot). Complex-scaling changes the orientation and size of the trapezoid. While the manifold distances \cite{chakraborty2019surreal} from the black dot to the colored dots are identical, the Division Layer output is invariant and distinct for each dot (the same colored dots around $0^\circ$). 
    }
    \label{fig:DivCinv}
\end{figure}
}

\def\figBigModelsNew#1{
\begin{figure}[#1]
    \centering
    \setlength{\tabcolsep}{1pt}
    \begin{tabular}{@{}c@{}}
    \begin{subfigure}[b]{5cm} \centering
    \small
    \begin{tabular}{crr}
    \toprule
    Method  & \#Params &  \%Acc 
    \\ \midrule
    DCN \cite{trabelsi2017deep} &  1.7M  &    92.8 
    \\ \midrule

    Ours &    1.7M  &   \textbf{93.7} 
    \\ \bottomrule

    \end{tabular}
    \small
      \caption{CIFAR10 LAB accuracy for large models} \label{fig:cifar10_result} \end{subfigure} \\
    \begin{subfigure}[b]{0.9\linewidth} \centering \includegraphics[width=0.9\linewidth]{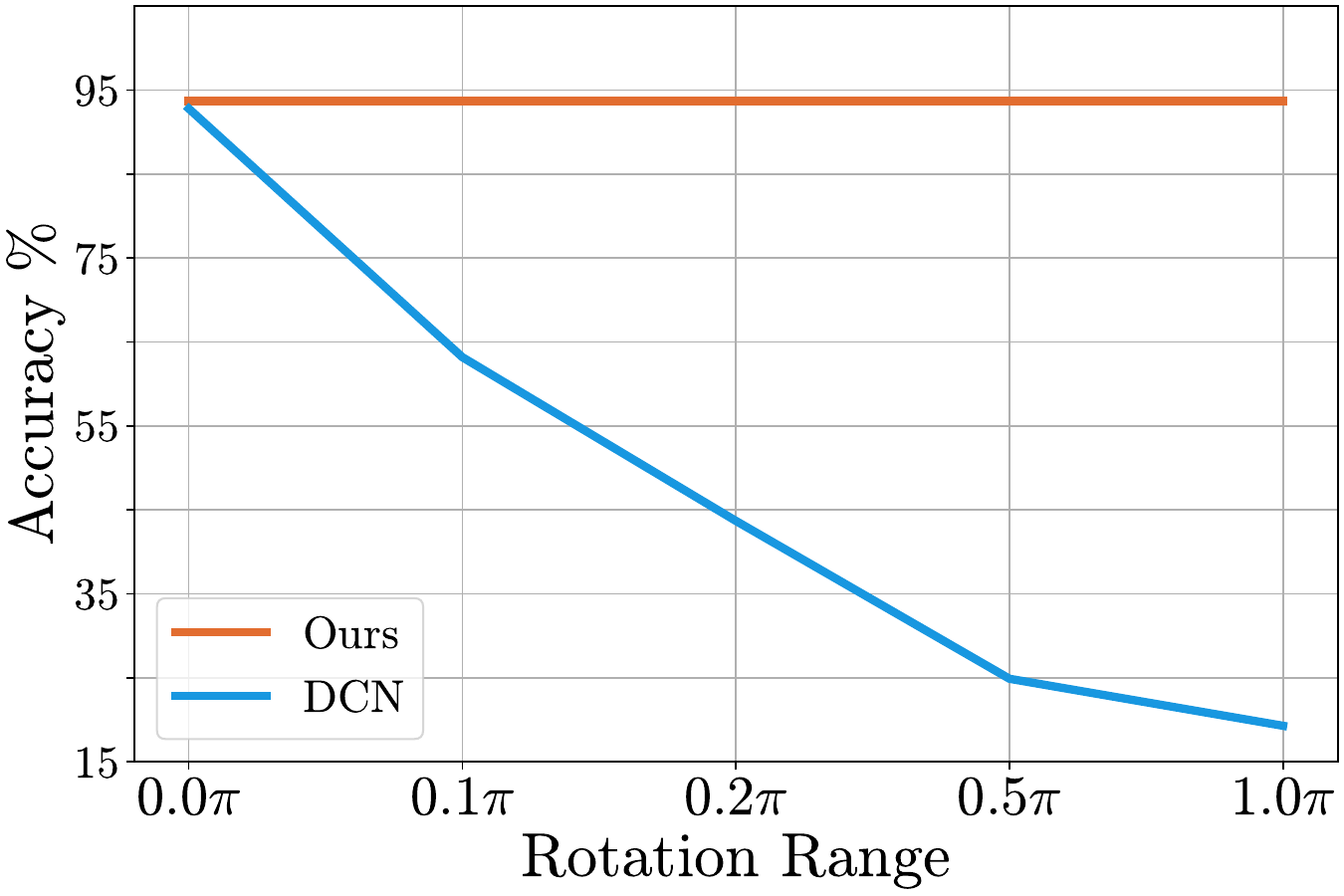}    \caption{Robustness Curve for large models} \label{fig:rotrob_new} \end{subfigure}    
    \end{tabular}
    \vspace*{-3mm}
    \caption{Our model beats DCN while additionally achieving complex-scale invariance. \textbf{a)} We train DCN and CDS on CIFAR10 with the LAB encoding, achieving higher accuracy. This result is consistent with (Table \ref{table:conv_compare_all}), with smaller margin due to the larger capacity of big models. \textbf{b)} DCN is susceptible to complex-valued scaling. Similar to Figure 1g, we plot average accuracy under different rotation ranges. DCN accuracy degrades under complex-valued scaling, while our method is robust.}
    \label{fig:rotrobbignew}
\end{figure}
}

\def\figClarification#1{
\begin{figure*}[#1]
    \centering
    \setlength{\tabcolsep}{1pt}
    \begin{tabular}{@{}ccc@{}}
    \begin{subfigure}[b]{0.3\textwidth} \centering \includegraphics[width=6cm]{fig/bias_var_bar.pdf}  \label{fig:biasvarrgb}  \caption{\textbf{RGB} encoding} \end{subfigure} &   
    \begin{subfigure}[b]{0.3\textwidth} \centering \includegraphics[width=6cm]{fig/bias_var_bar_lab.pdf}  \label{fig:biasvarlab}  \caption{\textbf{LAB} encoding} \end{subfigure} &  
    \begin{subfigure}[b]{0.3\textwidth} \centering \includegraphics[width=6cm]{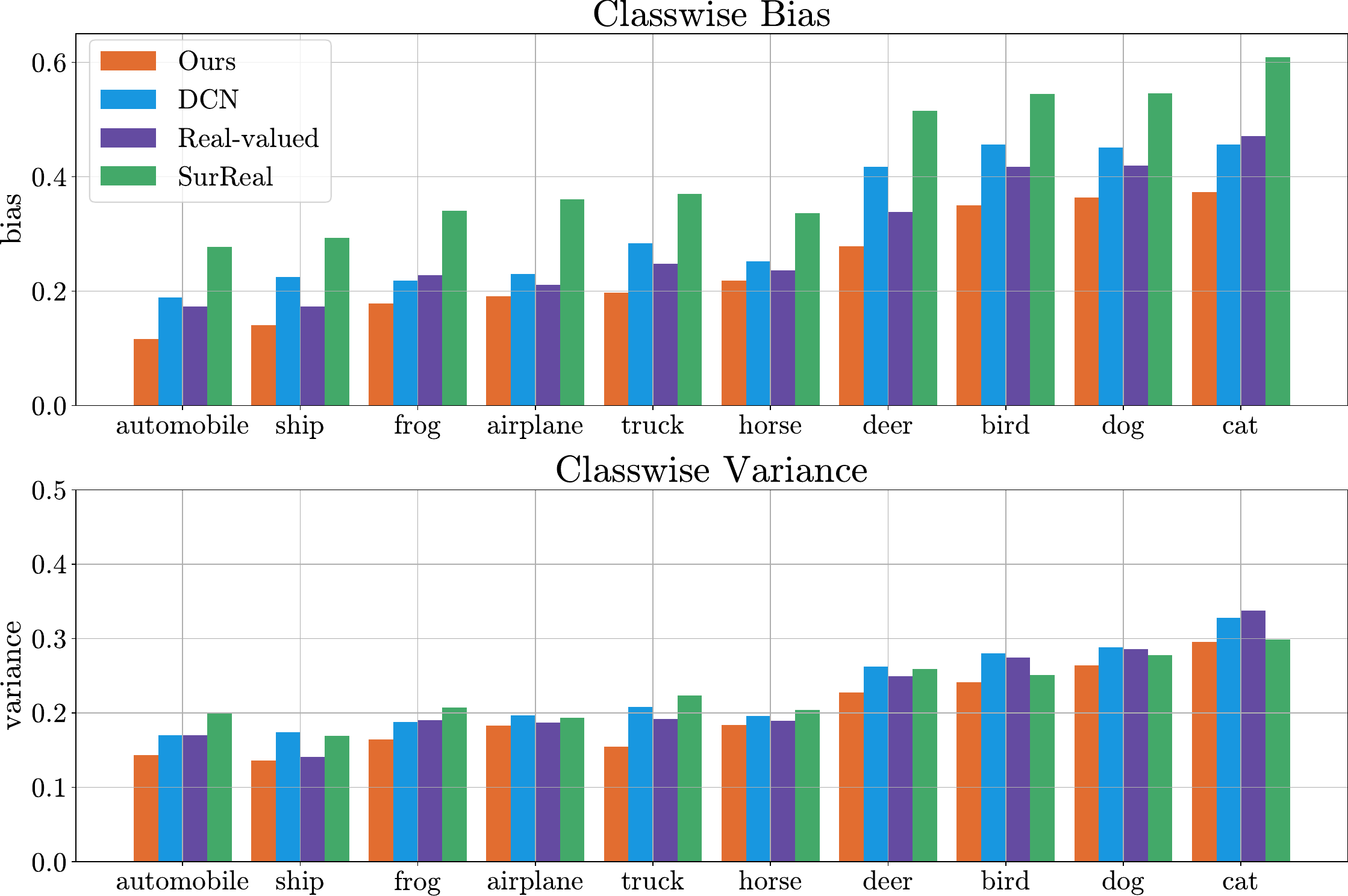}  \label{fig:biasvarsliding}  \caption{\textbf{Sliding} encoding} \end{subfigure}

    \end{tabular}
    \vspace*{-3mm}
    \caption{Our method demonstrates lowest overall bias and variance on CIFAR10. We followed \cite{wang2021longtailed} for CIFAR10 models with RGB encoding. Classes are ordered in ascending order of bias for our model. Our model consistently shows the lowest bias for each class in every encoding, and lowest variance in 8, 9, and 10 out of 10 classes with RGB, LAB, and Sliding encodings respectively.}
    \label{fig:clari}
\end{figure*}
}

%% file: table_defn.tex
\def\tabConvCompareOverall#1{
\begin{table*}[t]
\begin{center}
\small
\begin{tabular}{c|r|lll|lll|lll}
\toprule
Method                & \multicolumn{1}{c|}{\# Param} & \multicolumn{3}{c|}{CIFAR10}                                                     & \multicolumn{3}{c|}{CIFAR100}                                                    & \multicolumn{3}{c}{SVHN}                                                        \\
\midrule
\multicolumn{1}{l|}{} & \multicolumn{1}{l|}{}              & \multicolumn{1}{c}{RGB} & \multicolumn{1}{c}{LAB} & \multicolumn{1}{c|}{Sliding} & \multicolumn{1}{c}{RGB} & \multicolumn{1}{c}{LAB} & \multicolumn{1}{c|}{Sliding} & \multicolumn{1}{c}{RGB} & \multicolumn{1}{c}{LAB} & \multicolumn{1}{c}{Sliding} \\ 
\midrule
DCN \cite{trabelsi2017deep}                  & 66,858                              & 65.17                   & 58.64                   & 63.83                        & 32.52                   & 27.36                   & 28.87                        & 85.26                   & 84.43                   & 87.44                       \\
SurReal \cite{chakraborty2019surreal}               & 35,274                              & 50.68                   & 53.02                   & 54.61                        & 23.57                   & 25.97                   & 26.66                        & 80.51                   & 53.48                   & 80.79                       \\
Real-valued CNN       & 34,282                              & 64.43                   & 63                      & 63.43                        & 31.93                   & 31.72                   & 31.93                        & 87.47                   & 84.93                   & 87.37                       \\
\midrule
Ours (Type-I)      & 24,241                              & \textbf{69.23}          & 67.17                   & 68.7                         & 36.92                   & 37.81                   & 38.51                        & \textbf{89.39}          & \textbf{88.86}          & \textbf{90.25}              \\
Ours (Type-E)    & 25,745                              & 68.48                   & \textbf{67.58}          & \textbf{69.19}               & \textbf{41.83}          & \textbf{39.55}          & \textbf{42.08}               & 77.19                   & 74.21                   & 88.39                      \\
\bottomrule
\end{tabular}

\end{center}
\vspace*{-5mm}
\caption{Our models outperform the baselines' CIFARnet versions on real-valued datasets. \textit{Type-I} model performs better on easier datasets like SVHN, and \textit{Type-E} performs better on difficult datasets like CIFAR100. In contrast, SurReal does not scale to large datasets.}
\label{table:conv_compare_all}
\end{table*}
}

\def\tabSurRealCifarNet#1{
\begin{table*}[#1]\centering
\caption{SurReal CIFAR Model Architecture}
\setlength{\tabcolsep}{2pt}
\begin{tabular}{@{}lccccc@{}}
 \toprule
 {\bf Layer Type} & {\bf Input Shape} & {\bf Kernel} & {\bf Stride} & {\bf Padding}& {\bf Output Shape}\\
  \midrule
\bf Complex CONV&\([3, 32,32]\)	& $3\times 3$	&2  &1 &\([16,16,16]\)\\
\midrule
\bf $G$-transport&\([16,16,16]\)   & -         & -    &-   &\([16,16,16]\)\\
\midrule 
\bf Complex CONV&\([16,16,16]\)	& $3\times 3$	&2  &1 &\([32,8,8]\)\\
\midrule
\bf $G$-transport&\([32,8,8]\)   & -         & -   &-    &\([32,8,8]\)\\
\midrule 
\bf Complex CONV&\([32,8,8]\)	& $3\times 3$	&2  &1 &\([64,4,4]\)\\

\midrule 
\bf $G$-transport&\([64,4,4]\)   & -         & -   &-    &\([64,4,4]\)\\
\midrule 
\bf Distance   Layer&\([64,4,4]\)	& -	&-  &- &\([64,4,4]\)\\

\midrule 
\bf Average Pooling &\([64,4,4]\)	& $4\times 4$	&-  &-  &\([64,1,1]\)\\

\midrule 
\bf FC       &\([64]\)           & -         &-      &- &\([128]\)      \\
\midrule 
\bf ReLU       &\([128]\)           & -         &-      &- &\([128]\)      \\
\midrule 
\bf FC       &\([128]\)           & -         &-      &- &\([10]\)      \\
  \bottomrule
 \end{tabular}
\label{tab:SurRealCifarNet} 
\end{table*}
}

\def\tabDCNCifarNet#1{
\begin{table*}[#1]\centering
\caption{DCN CIFAR Model Architecture}
\setlength{\tabcolsep}{2pt}
\begin{tabular}{@{}lccccc@{}}
 \toprule
 {\bf Layer Type} & {\bf Input Shape} & {\bf Kernel} & {\bf Stride} & {\bf Padding}& {\bf Output Shape}\\
  \midrule
\bf Complex CONV&\([3, 32,32]\)	& $3\times 3$	&2  &1 &\([16,16,16]\)\\
\midrule
\bf \crelu &\([16,16,16]\)   & -         & -    &-   &\([16,16,16]\)\\
\midrule 
\bf Complex CONV&\([16,16,16]\)	& $3\times 3$	&2  &1 &\([32,8,8]\)\\
\midrule
\bf \crelu &\([32,8,8]\)   & -         & -   &-    &\([32,8,8]\)\\
\midrule 
\bf Complex CONV&\([32,8,8]\)	& $3\times 3$	&2  &1 &\([64,4,4]\)\\

\midrule 
\bf \crelu &\([64,4,4]\)   & -         & -   &-    &\([64,4,4]\)\\
\midrule 
\bf Average Pooling &\([64,4,4]\)	& $4\times 4$	&-  &-  &\([64,1,1]\)\\
\midrule 
\bf Complex-to-Real &\([64,1,1]\)	& $4\times 4$	&-  &-  &\([128]\)\\

\midrule 
\bf FC       &\([128]\)           & -         &-      &- &\([128]\)      \\
\midrule
\bf ReLU       &\([128]\)           & -         &-      &- &\([128]\)      \\
\midrule 
\bf FC       &\([128]\)           & -         &-      &- &\([10]\)      \\
  \bottomrule
 \end{tabular}
\label{tab:DCNCifarNet} 
\end{table*}
}

\def\tabRealCifarNet#1{
\begin{table*}[#1]\centering
\caption{2-Channel Real-Valued CIFAR Model Architecture}
\setlength{\tabcolsep}{2pt}
\begin{tabular}{@{}lccccc@{}}
 \toprule
 {\bf Layer Type} & {\bf Input Shape} & {\bf Kernel} & {\bf Stride} & {\bf Padding}& {\bf Output Shape}\\
  \midrule
\bf CONV&\([3, 32,32]\)	& $3\times 3$	&2  &1 &\([16,16,16]\)\\
\midrule
\bf ReLU &\([16,16,16]\)   & -         & -    &-   &\([16,16,16]\)\\
\midrule 
\bf CONV&\([16,16,16]\)	& $3\times 3$	&2  &1 &\([32,8,8]\)\\
\midrule
\bf ReLU &\([32,8,8]\)   & -         & -   &-    &\([32,8,8]\)\\
\midrule 
\bf CONV&\([32,8,8]\)	& $3\times 3$	&2  &1 &\([64,4,4]\)\\
\midrule 
\bf ReLU&\([64,4,4]\)   & -         & -   &-    &\([64,4,4]\)\\
\midrule 
\bf Average Pooling &\([64,4,4]\)	& $4\times 4$	&-  &-  &\([64,1,1]\)\\
\midrule 
\bf FC       &\([64]\)           & -         &-      &- &\([128]\)      \\
\midrule 
\bf ReLU       &\([128]\)           & -         &-      &- &\([128]\)      \\
\midrule 
\bf FC       &\([128]\)           & -         &-      &- &\([10]\)      \\
  \bottomrule
 \end{tabular}
\label{tab:RealCifarNet} 
\end{table*}
}

\def\tabDCNmstar#1{
\begin{table*}[#1]\centering
\caption{DCN Down-sampling Block for MSTAR}
\setlength{\tabcolsep}{2pt}
\begin{tabular}{@{}lccccc@{}}
 \toprule
 {\bf Layer Type} & {\bf Input Shape} & {\bf Kernel} & {\bf Stride} & {\bf Padding}& {\bf Output Shape}\\
  \midrule
\bf Complex CONV&\([1, 128, 128]\)	& $3\times 3$	&2  &1 &\([12,64,64]\)\\
\midrule
\bf ComplexBatchNorm &\([12,64,64]\)   & -         & -    &-   &\([12,64,64]\)\\
\midrule 
\bf Complex CONV&\([1, 64, 64]\)	& $3\times 3$	&2  &1 &\([12,32,32]\)\\
\midrule
\bf ComplexBatchNorm &\([12,32,32]\)   & -         & -    &-   &\([12,32,32]\)\\
  \bottomrule
 \end{tabular}
\label{tab:DCNmstar} 
\end{table*}
}

\def\tabConvCompare#1{
\begin{table}[t]
\begin{center}
\small
\scalebox{0.9}{
\begin{tabular}{|l|c|c|c|c|c|c|c|c|c|}
\hline
 & \multicolumn{3}{|c|}{CIFAR10} & \multicolumn{3}{|c|}{CIFAR100} & \multicolumn{3}{|c|}{SVHN}

\\ \hline

 Method &  RGB & LAB & Orientation  & RGB & LAB & Orientation   & RGB & LAB & Orientation
\\ \hline

Ours     &   &  &  & & $28.14$ & & $86.77$  & $86.17$ & 
\\ \hline 

Constrained + G &   $61.23$   & $56.37$  & $57.31$& $30.68$ & $29.28$ & $28.15$& $82.33$ & $81.33$ & $86.08$   
\\ \hline

Constrained &      &   &  & $22.26$ & $22.82$ & & $75.65$ & $74.52$&   
\\ \hline 

\end{tabular}
}
\end{center}
\caption{Comparison between our model and surreal constraint-enforced model. In this constrained model, we utilize the same architecture as our model, but replace complex convolutions with those defined in SurReal, where kernel weights are real, positive and sum to one. These experiments show that our complex-valued weights generalize better than SurReal weighted Frechet mean.}
\label{table:conv_compare}
\end{table}
}

\def\tabMstar#1{
\begin{table}[#1]
\begin{center}
\small
\scalebox{0.95}{

\begin{tabular}{c|r|ccccc}
\toprule
Model              & \# Params & 5\%  & 10\% & 50\% & 90\% & 100\%         \\ \midrule
Real        & 33,050    & 47.4 & 46.6 & 60.6 & 73   & 66.9          \\ \midrule
SurReal \cite{chakraborty2019surreal}            & 63,690    & 61.1 & 68   & 90.3 & \textbf{95.6} & 94.9          \\ \midrule
DCN \cite{trabelsi2017deep}                & 863,587   & 49.8 &   47   & 81.9 &  89.1 & 89.1          \\ \midrule
Ours            & 29,536    & \textbf{69.5} & \textbf{78.3} & \textbf{91.3} & 95.2 & \textbf{96.1}          \\ \bottomrule
\end{tabular}
}
\end{center}
\vspace*{-5mm}
\caption{Our method achieves the best accuracy and generalization with the fewest parameters.  We report accuracy on varying proportions of MSTAR training data. The performance gap is wider for smaller train-sets, with Real-CNN and DCN failing to generalize.}
\label{table:mstar_result}
\end{table}
}

\def\tabCifarResult#1{
\begin{table}[#1]
\begin{center}
\scalebox{1}{
\small
\begin{tabular}{crr}
\toprule
Method  & \#Params &  \%Acc 
\\ \midrule
DCN \cite{trabelsi2017deep} &  1.7M  &    92.8 
\\ \midrule

Ours &    1.7M  &   \textbf{93.7} 
\\ \bottomrule

\end{tabular}
}
\end{center}
\small
\vspace*{-5mm}
\caption{Our model beats DCN along with added complex-scale invariance. We train DCN and CDS on CIFAR10 with the LAB encoding. This result is consistent with (Table \ref{table:conv_compare_all}), with smaller margin due to the larger capacity of big models.}
\label{table:cifar10_result}
\end{table}
}

\def\tabrealmstar#1{
\begin{table*}[#1]\centering
\caption{MSTAR Real-valued Model Architecture}
\setlength{\tabcolsep}{2pt}
\begin{tabular}{@{}lcccc@{}}
 \toprule
 {\bf Layer Type} & {\bf Input Shape} & {\bf Kernel} & {\bf Stride} & {\bf Output Shape}\\
  \midrule

\bf  CONV &\([2,100,100]\)     &$5\times 5$     &1      &\([30,96,96]\)\\
\midrule 
\bf GroupNorm$+$ReLU&\([30,96,96]\)     & -         & -         &\([30,96,96]\)\\
\midrule 
\bf ResBlock          &\([30,96,96]\)        & -    & -      &\([40,96,96]\)  \\
\midrule 
\bf MaxPool           &\([40,96,96]\)     &$2\times 2$     &2    &\([40,48,48]\)  \\
\midrule 
\bf CONV            &\([40,48,48]\)     &$5\times 5$     &3     &\([50,15, 15]\)  \\
\midrule 
\bf GroupNorm$+$ReLU&\([50,15, 15]\)       & -         & -         &\([50,15,15]\)  \\
\midrule 
\bf ResBlock&\([50,15,15]\)       & -         & -          &\([60,15,15]\)  \\
\midrule 
\bf CONV         &\([60,15,15]\)       &$2\times 2$     &1    &\([70,14,14]\)  \\
\midrule 
\bf GroupNorm$+$ReLU&\([70,14,14]\)       & -         & -          &\([70,14,14]\)  \\
\midrule 
\bf AveragePool          &\([70,14,14]\)           & -         & -         &\([70]\)      \\
\midrule 
\bf FC           &\([70]\)           & -         & -         &\([30]\)      \\
\midrule 
\bf ReLU            &\([30]\)           & -         & -        &\([30]\)      \\
\midrule 
\bf FC            &\([30]\)           & -         & -      &\([10]\)      \\

  \bottomrule
 \end{tabular}
\label{tab:realmstar} 
\end{table*}
}

\def\tabAblations#1{
\begin{table}[t]
\begin{center}
\small
\begin{tabular}{c|rrr}
\toprule
Method                &  Accuracy                                                  \\
\midrule
Division Layer            & \textbf{67.17}     \\
Conjugate Layer      & 66.73      \\
\midrule
Euclidean Distance    & 67.17     \\
Manifold Distance     & \textbf{68.54}      \\
\midrule
GTReLU (r=0)     & 67.17     \\
GTReLU (r=0.1)     & \textbf{68.14}     \\
GTReLU (r=1)     & 49.15     \\
\bottomrule
\end{tabular}

\end{center}
\vspace*{-5mm}
\caption{Ablation test results for our Type-I model on CIFAR 10 with LAB encoding. We find that Division Layer, Manifold Distance, and a GTReLU threshold of $r=0.1$ perform the best.}
\label{table:ablations}
\end{table}
}

\def\tabWFM#1{
\begin{table}[t]
\begin{center}
\small
\scalebox{0.9}{
\begin{tabular}{c|r|rrr}
\toprule
Layer Type   & Params             &  Acc (\%)                                                   \\
\midrule
Complex     & 66,858   &    58.6 \\
\midrule
Real    & 34,282   &   \textbf{63.4}  \\
\midrule
wFM & 42,154   &     52.2 \\
\bottomrule
\end{tabular}

}
\end{center}
\vspace*{-5mm}
\caption{wFM results in significant reductions in accuracy. We train models on CIFAR10 with LAB encoding and tabulate the resulting test accuracy. Compared to a real-valued CNN, wFM results in significantly lower accuracy due to its restricted formulation.}
\label{table:WFM}
\end{table}
}

\def\tabourECifarNet#1{
\begin{table*}[#1]\centering
\caption{Our (Type-E) CIFAR Model Architecture}
\setlength{\tabcolsep}{2pt}
\begin{tabular}{@{}lccccc@{}}
 \toprule
 {\bf Layer Type} & {\bf Input Shape} & {\bf Kernel} & {\bf Stride} & {\bf Padding}& {\bf Output Shape}\\
  \midrule
\bf Econv        &\([3, 32,32]\)	& $3\times 3$	&2  &1 &\([16,16,16]\)\\
\midrule
\bf Eq. GTReLU       &\([16,16,16]\)   & -         & -    &-   &\([16,16,16]\)\\
\midrule 
\bf Econv        &\([16,16,16]\)	& $3\times 3$	&2  &1 &\([32,8,8]\)\\
\midrule
\bf Eq. GTReLU       &\([32,8,8]\)   & -         & -   &-    &\([32,8,8]\)\\
\midrule 
\bf Econv        &\([32,8,8]\)	& $3\times 3$	&2  &1 &\([64,4,4]\)\\

\midrule 
\bf Eq. GTReLU       &\([64,4,4]\)   & -         & -   &-    &\([64,4,4]\)\\
\midrule 
\bf Average Pooling       &\([64,4,4]\)	& $4\times 4$	&-  &-  &\([64,1,1]\)\\
\midrule 
\bf Equivariant FC        &\([64]\)           & -         &-      &- &\([128]\)      \\
\midrule
\bf Invariant Prototype Distance       &\([128]\)           & -         &-      &- &\([10]\)      \\
  \bottomrule
 \end{tabular}
\label{tab:ourECifarNet} 
\end{table*}
}

\def\tabourICifarNet#1{
\begin{table*}[#1]\centering
\caption{Our (Type-I) CIFAR Model Architecture}
\setlength{\tabcolsep}{2pt}
\begin{tabular}{@{}lccccc@{}}
 \toprule
 {\bf Layer Type} & {\bf Input Shape} & {\bf Kernel} & {\bf Stride} & {\bf Padding}& {\bf Output Shape}\\
  \midrule
\bf Econv        &\([3, 32,32]\)	& $3\times 3$	&2  &1 &\([16,16,16]\)\\
\midrule
\bf Division Layer       &\([16,16,16]\)   & $3\times 3$         & -    & 1   &\([16,16,16]\)\\
\midrule
\bf GTReLU       &\([16,16,16]\)   & -         & -    &-   &\([16,16,16]\)\\
\midrule 
\bf Econv        &\([16,16,16]\)	& $3\times 3$	&2  &1 &\([32,8,8]\)\\
\midrule
\bf GTReLU       &\([32,8,8]\)   & -         & -   &-    &\([32,8,8]\)\\
\midrule 
\bf Econv        &\([32,8,8]\)	& $3\times 3$	&2  &1 &\([64,4,4]\)\\

\midrule 
\bf GTReLU       &\([64,4,4]\)   & -         & -   &-    &\([64,4,4]\)\\
\midrule 
\bf Average Pooling       &\([64,4,4]\)	& $4\times 4$	&-  &-  &\([64,1,1]\)\\
\midrule 
\bf Equivariant FC        &\([64]\)           & -         &-      &- &\([128]\)      \\
\midrule
\bf Prototype Distance       &\([128]\)           & -         &-      &- &\([10]\)      \\
  \bottomrule
 \end{tabular}
\label{tab:ourICifarNet} 
\end{table*}
}

\def\tabourBIG#1{
\begin{table*}[#1]\centering
\caption{Our CDS-Large Model Architecture}
\setlength{\tabcolsep}{2pt}
\begin{tabular}{@{}lccccc@{}}
 \toprule
 {\bf Layer Type} & {\bf Input Shape} & {\bf Kernel} & {\bf Stride} & {\bf Padding}& {\bf Output Shape}\\
  \midrule
\bf Econv        &\([3, 32,32]\)	& $3\times 3$	&1  &1 &\([64, 32, 32]\)\\
\midrule
\bf Conjugate Layer       &\([64,32,32]\)   & $1\times 1$         & -    &  -   &\([64,32,32]\)\\
\midrule
\bf Econv (Groups=2)        &\([64, 32,32]\)	& $3\times 3$	&1  &1 &\([64, 32, 32]\)\\
\midrule
\bf ComplexBatchNorm        &\([64, 32,32]\)	& -	& -  & - &\([64, 32, 32]\)\\
\midrule
\bf \crelu        &\([64, 32,32]\)	& -	& -  &- &\([64, 32, 32]\)\\
\midrule
\bf Econv (Groups=2)        &\([64, 32,32]\)	& $3\times 3$	&1  &1 &\([128, 32, 32]\)\\
\midrule
\bf ComplexBatchNorm        &\([128, 32,32]\)	& -	& -  & - &\([128, 32, 32]\)\\
\midrule
\bf \crelu        &\([128, 32,32]\)	& -	& -  &- &\([128, 32, 32]\)\\
\midrule
\textbf{Eq. MaxPool}      &\([128, 32,32]\)	& $2 \times 2 $	& -  &- &\([128, 16, 16]\)\\
\midrule
\textbf{ResBlock(groups=2)}      &\([128,16,16]\)	& -	& -  &- &\([128, 16, 16]\)\\
\midrule
\bf Econv (Groups=4)        &\([128,16,16]\)	& $3\times 3$	&1  &1 &\([256,16,16]\)\\
\midrule
\bf ComplexBatchNorm        &\([256,16,16]\)	& -	& -  & - &\([256,16,16]\)\\
\midrule
\bf \crelu        &\([256,16,16]\)	& -	& -  &- &\([256,16,16]\)\\
\midrule
\textbf{Eq. MaxPool }     &\([256,16,16]\)	& $2 \times 2 $	& -  &- &\([256,8,8]\)\\
\midrule
\bf Econv (Groups=2)        &\([256,8,8]\)	& $3\times 3$	&1  &1 &\([512,8,8]\)\\
\midrule
\bf ComplexBatchNorm        &\([512,8,8]\)	& -	& -  & - &\([512,8,8]\)\\
\midrule
\bf \crelu        &\([512,8,8]\)	& -	& -  &- &\([512,8,8]\)\\
\midrule
\textbf{Eq. MaxPool }     &\([512,8,8]\)	& $2 \times 2 $	& -  &- &\([512,4,4]\)\\
\midrule
\textbf{ResBlock(groups=4)}      &\([512,4,4]\)	& -	& -  &- &\([512,4,4]\)\\
\midrule
\textbf{Eq. MaxPool }     &\([512,4,4]\)	& $2 \times 2 $	& -  &- &\([512,1,1]\)\\
\midrule
\textbf{Fully Connected }     &\([1024]\)	& -	& -  &- &\([10]\)\\
  \bottomrule
 \end{tabular}
\label{tab:ourBIG} 
\end{table*}
}

\def\tabourMSTAR#1{
\begin{table*}[#1]\centering
\caption{Our MSTAR Model Architecture}
\setlength{\tabcolsep}{2pt}
\begin{tabular}{@{}lccccc@{}}
 \toprule
 {\bf Layer Type} & {\bf Input Shape} & {\bf Kernel} & {\bf Stride} & {\bf Padding}& {\bf Output Shape}\\
  \midrule
\bf Econv (Groups=5)            &\([1, 100,100]\)	& $5\times 5$	&1  &0 &\([5, 96, 96]\)\\
\midrule
\bf Eq. GTReLU       &\([5, 96, 96]\)   & -         & -    &-   &\([5, 96, 96]\)\\
\midrule
\bf Eq. MaxPool      &\([5, 96, 96]\)   &$2\times 2$     &2   &-   &\([5, 48, 48]\)\\
  \midrule
\bf Econv            &\([5, 48, 48]\)	& $3\times 3$	&2  &0 &\([5, 23, 23]\)\\
\midrule
\bf Eq. GTReLU       &\([5, 23, 23]\)   & -         & -    &-   &\([5, 23, 23]\)\\
\midrule
\bf Division Layer   &\([5, 23, 23]\)   & $3\times 3$         & -    &-   &\([5, 21, 21]\)\\
\midrule 
\bf  Complex-to-Real &\([5,21,21]\)     &-     &-    & -  &\([15,21,21]\)\\
\midrule 
\bf  CONV (Groups=5) &\([15,21,21]\)     &$5\times 5$     &1    & -  &\([30,17,17]\)\\
\midrule 
\bf GroupNorm$+$ReLU&\([30,17,17]\)     & -         & -       & -   &\([30,17, 17]\)\\
\midrule 
\bf ResBlock          &\([30,17,17]\)     & -     & -    & -  &\([40,17,17]\)  \\
\midrule 
\bf MaxPool           &\([40,17,17]\)     &$2\times 2$     &2   & -   &\([40,8,8]\)  \\
\midrule 
\bf CONV (Groups=5)            &\([40,8,8]\)     &$5\times 5$     &3    & -  &\([50,2,2]\)  \\
\midrule 
\bf GroupNorm$+$ReLU&\([50,2,2]\)       & -         & -      & -    &\([50,2,2]\)  \\
\midrule 
\bf ResBlock&\([50,2,2]\)       & -         & -      & -    &\([60,2,2]\)  \\
\midrule 
\bf CONV (Groups=5)            &\([60,2,2]\)       &$2\times 2$     &1  & -   &\([70,1,1]\)  \\
\midrule 
\bf GroupNorm$+$ReLU&\([70,1,1]\)       & -         & -      & -      &\([70,1,1]\)  \\
\midrule 
\bf FC           &\([70]\)           & -         & -      & -    &\([30]\)      \\
\midrule 
\bf ReLU            &\([30]\)           & -         & -     & -     &\([30]\)      \\
\midrule 
\bf FC            &\([30]\)           & -         & -      & -    &\([10]\)      \\
  \bottomrule
 \end{tabular}
\label{tab:ourMSTAR} 
\end{table*}
}

%% file: 0abs.tex
\begin{abstract}

We study complex-valued scaling as a type of symmetry natural and unique to complex-valued measurements and representations. Deep Complex Networks (DCN) extends real-valued algebra to the complex domain without addressing complex-valued scaling.  SurReal takes a restrictive manifold view of complex numbers, adopting a distance metric to achieve complex-scaling invariance while losing rich complex-valued information.  

We analyze complex-valued scaling as a co-domain transformation and design novel equivariant and invariant neural network layer functions for this special transformation.  We also propose novel complex-valued representations of RGB images, where complex-valued scaling indicates hue shift or correlated changes across color channels.

Benchmarked on MSTAR, CIFAR10, CIFAR100, and SVHN, our co-domain symmetric (CDS) classifiers deliver higher accuracy,  better generalization, robustness to co-domain transformations,  and lower model bias and variance than DCN and SurReal with far fewer parameters.

\end{abstract}

%% file: 1intro.tex
\section{Introduction}
Symmetry is one of the most powerful tools in the deep learning repertoire. Naturally occurring symmetries lead to structured variation in natural data, and so modeling these symmetries greatly simplifies learning \cite{weiler2018steerable}.  A key factor behind the success of Convolutional Neural Networks~\cite{cnn_lecun} is their ability to capture the \textit{translational symmetry} of image data. Similarly, PointNet~\cite{qi2017pointnet} captures \textit{permutation symmetry} of point-clouds. These symmetries are formalized as \textit{invariance} or \textit{equivariance} to a group of transformations \cite{eade2014lie}.  This view is taken to model a general class of \textit{spatial symmetries}, including 2D/3D rotations \cite{lieconv, cohen2016group, vn2021}.  However, this line of research has primarily focused on real-valued data.

We explore complex-valued data which arise naturally in \textbf{1)} remote sensing such as synthetic aperture radar (SAR), medical imaging such as magnetic resonance imaging (MRI), and radio frequency communications; \textbf{2)} spectral representations of real-valued data such as Fourier Transform \cite{Mallat2016, Kondor-icml18}; and \textbf{3)} physics and engineering applications \cite{complexBk1998}. In deep learning, complex-valued models have shown several benefits over their real-valued counterparts: larger representational capacity \cite{nitta2003xor}, more robust embedding \cite{stella_ae} and associative memory \cite{pmlr-v48-danihelka16}, more efficient multi-task learning \cite{superpos}, and higher quality MRI image reconstruction \cite{mrirecon}. We approach complex-valued deep learning from a symmetry perspective: {\it Which symmetries are inherent in complex-valued data, and how do we exploit them in modeling?}

\figCoDomain{!t}{0.48}
\figInvariance{!t}

One type of symmetry inherent to complex-valued data is \textit{complex-valued scaling ambiguity} \cite{chakraborty2019surreal}. For example, consider a complex-valued MRI or SAR signal ${\bf z}$. Due to the nature of signal acquisition, $\textbf{z}$ could be subject to global magnitude scaling and phase offset represented by a complex-valued scalar $s$, thus becoming $s \cdot {\bf z}$.

A complex-valued classifier takes input ${\bf z}$ and ideally should focus on discriminating among instances from different classes, not on the instance-wise variation  $s\cdot {\bf z}$ caused by complex-valued scaling.  Formally, function $f$ is called \textbf{complex-scale invariant} 
if $ f(s\cdot\mathbf{z}) = f(\mathbf{z})$ and called \textbf{complex-scale equivariant} if $f(s\cdot \mathbf{z}) = s\cdot f(\mathbf{z})$.

We distinguish two types of image transformations, viewing an image as a function defined over spatial locations.  Complex-valued scaling of a complex-valued image is a transformation in the co-domain of the image function, as opposed to a spatial transformation in the domain of the image (Figure \ref{fig:co_domain}). Formally, $I: \mathbb{R}^D \to \mathbb{C}^K$ denotes a complex-valued image of $K$ channels in the $D$-dimensional space, where  
 $\mathbb{R}$ ($\mathbb{C}$) denotes the set of real (complex) numbers.  Some common $(D,K)$ are (2,1) for grayscale images, (2,3) for RGB, and $(3,6+)$ for diffusion tensor images.
 \ol{enumerate}{0}{
\item{\bf Domain transformation} $T: \mathbb{R}^D\!\to\!\mathbb{R}^D$ transforms the spatial coordinates of an image, resulting in a spatially warped image $I(T(x))$, $x\in \mathbb{R}^D$. Translation, rotation, and scaling are examples of domain transformations. 
\item{\bf Co-domain transformation} $T': \mathbb{C}^K \to \mathbb{C}^K$ maps the pixel value to another, resulting in a color adjusted image  $T'(I(x))$, $x\in \mathbb{R}^D$. Complex-valued scaling and color distortions are examples of co-domain transformations.
}
Complex-valued scaling thus presents not only a practical setting but also a case study for co-domain transformations.  

Existing complex-valued deep learning methods such as DCN \cite{trabelsi2017deep} are sensitive to complex-valued scaling (Fig \ref{fig:dcn_tsne}). A pre-processing trick to remove such scaling ambiguity is to simply normalize all the pixel values by setting their average phase to $0$ and magnitude to $1$, but this process introduces artifacts when the phase distribution varies greatly with the content of the image (Figure \ref{fig:PN}). SurReal \cite{chakraborty2019surreal} applies manifold-valued deep learning to complex-valued data, but this framework only captures the manifold aspect and not the complex algebra of complex-valued data. Thus, a more general, principled method is needed. We propose novel layer functions for complex-valued deep learning by studying how they preserve co-domain symmetry. Specifically, we study whether each layer-wise transformation achieves equivariance or invariance to complex-valued scaling.

\textbf{Our contributions:} {\bf 1)} We derive complex-scaling equivariant and invariant versions of common layers used in computer vision pipelines. Our model circumvents the limitations of SurReal \cite{chakraborty2019surreal} and scales to larger models and datasets. {\bf 2)} Our experiments on MSTAR, CIFAR 10, CIFAR 100, and SVHN datasets demonstrate a significant gain in generalization and robustness. {\bf 3)} We introduce novel complex-valued encodings of color, demonstrating the utility of using complex-valued representations for real-valued data. Complex-scaling invariance under our \textit{LAB} encoding automatically leads to color distortion robustness without the need for color jitter augmentation.

%% file: 2related.tex
\section{Related Work}

\textbf{Complex-Valued Processing:} Complex numbers are ubiquitous in mathematics, physics, and engineering \cite{oppenheim1999discrete, complexBk1998, mathews1970mathematical}. Traditional complex-valued data analysis involves higher-order statistics \cite{Kinsner2010, reichert1992automatic}. \cite{nitta2003xor} demonstrates higher representational capacity of complex-valued processing on the XOR problem. \cite{cadieu2012learning} proposes a sparse coding layer utilizing complex basis functions.  \cite{reichert2013neuronal} proposes a biologically meaningful complex-valued model. \cite{yu:bright09,yu:ae12} encodes data features in a complex vector and learns a metric for this embedding. \cite{mrirecon} applies complex-valued neural networks to MRI image reconstruction. \cite{nitta2002critical} investigates the role of critical points in complex neural networks. \cite{hirose2012generalization} demonstrates that complex networks have smaller generalization error than real-valued, and offers an overview of their convergence and stability. We refer the reader to \cite{trabelsi2017deep} for a more detailed account of complex-valued deep learning. 

\textbf{Transformation Equivariance/Invariance}: An important line of work \cite{lieconv, cohen2016group, worrall2019deep, marcos2017rotation} aims to develop convolutional layers equivariant to domain transformations like rotation/scaling. \cite{cohen2016group} introduces a principled method for producing group-equivariant layers for finite groups. \cite{lieconv} extends this work to Lie groups on continuous data. \cite{worrall2017harmonic} uses circular harmonics to produce deep neural networks equivariant to rotation and translation. \cite{cohen2018general} attempts to produce a general theory of group-equivariant CNNs on the Euclidean space and the sphere. \cite{cohen2019gauge} further extends the framework to local gauge transformations on the manifold.
This class of methods is well-suited for domain transformations. However, complex-valued scaling is a \textit{co-domain} transformation, and these methods are not applicable. \cite{vn2021} generalizes neurons to $\mathbb{R}^3$ vectors with 3D rotations as a co-domain transformation, introducing rotation-equivariant layers for point-clouds. In contrast, our method handles both the complex algebra and the geometry of complex-valued scaling.

\textbf{Complex-Valued Scaling}: Despite the increasing research interest in complex-valued neural networks, the problem of effectively dealing with complex-scaling ambiguity remains open. \cite{trabelsi2017deep, zhang2017complex, pat:complex17,patThesis2018} propose an extension of real neural architectures to the complex field by redefining basic building blocks such as complex convolution, batch normalization, and non-linear activation functions. However, these methods are not robust against complex-valued scaling. SurReal \cite{chakraborty2019surreal} tackles the problem of complex-scale invariance by adopting a manifold-based view of complex numbers. It models each complex number as an element of a manifold where complex-scaling corresponds to translation and uses tools from manifold-valued learning to create complex-scale invariant models. This results in better generalization to unseen complex-valued data with leaner models. However, the SurReal framework is restrictive (Sec. \ref{sec:eqconv}), and the complex-valued processing is forced to be linear (Sec. \ref{sec: eqnl}), preventing SurReal from scaling to large datasets (Table \ref{table:conv_compare_all}). 

%% file: 3model.tex
\section{Equivariance and Invariance for Complex-Valued Deep Learning}

\subsection{Equivariant Convolution}
\label{sec:eqconv}
In contrast to domain transformations which group-specific convolution layers for equivariance \cite{lieconv, cohen2016group}, any linear layer is equivariant to complex-valued scaling: For a linear function $L: \mathbb{C}^n \to \mathbb{C}^n$ with an input vector $x \in \mathbb{C}^n$ and complex scalar $s \in \mathbb{C}$, $L(s \cdot x) = s \cdot L(x)$. While a bias term is useful in real-valued neural networks, it turns the convolution layer into an affine function, destroying complex-scale equivariance. Thus, we remove the bias term from the complex-valued convolution used in DCN \cite{trabelsi2017deep}, restoring its equivariance. Additionally, we use Gauss' multiplication trick to speed up the convolution by $25\%$.

Given a complex-valued input feature map $\mathbf{z} = \mathbf{a}+i\mathbf{b} \in \mathbb{C}^{C \times H \times W}$ with $C$ channels and $H \times W$ pixels, and given a convolutional filter of size $K \times K$ with weight $\mathbf{W} = \mathbf{X}+i\mathbf{Y} \in \mathbb{C}^{C \times C \times K \times K}$, we define the Complex-Scale Equivariant Convolution as: \
\begin{align*}
     \operatorname{Econv}(\mathbf{z} ; \mathbf{W}) = \mathbf{W} * \mathbf{z}
    =& \ (\mathbf{X}+i\mathbf{Y}) * (\mathbf{a} + i \mathbf{b} ) \\
    =& \ \mathbf{t_1} - \mathbf{t_2} + i (\mathbf{t_3} - \mathbf{t_1} - \mathbf{t_2})
\end{align*}
where $\mathbf{t_1} = \mathbf{X}\ *\ \mathbf{a}$, $\mathbf{t_2} = \mathbf{Y}\ *\ \mathbf{b}$, $\mathbf{t_3} = (\mathbf{X}+\mathbf{Y})\ *\ (\mathbf{a}+\mathbf{b})$, and $\mathbf{X}*\mathbf{a}$ represents the convolution operation on $\mathbf{a}$ with weight $\mathbf{X}$. In contrast, SurReal uses weighted Frechet Mean (wFM), a restricted convolution where the weights are constrained to be real-valued, positive, and to sum to $1$. This restrictive formulation drastically reduces accuracy. 

\subsection{Equivariant Non-Linearity}
\label{sec: eqnl}
Non-linear activation functions such as ReLU are necessary to construct deep hierarchical representations. For complex-valued data, several alternatives have been studied \cite{trabelsi2017deep, patThesis2018, nonlin, chakraborty2019surreal}. \crelu, the most prominent example, computes ReLU independently on the real and imaginary parts of the input. Tangent ReLU (TReLU) \cite{chakraborty2019surreal} uses the polar representation, thresholding both magnitude and phase. 

However, these non-linearities are not complex-scale equivariant. DCN \cite{trabelsi2017deep} uses \crelu, failing to be robust against complex-scaling. SurReal does not use non-linearities in its complex-valued stages (see Tables I \& II in \cite{chakraborty2019surreal}). SurReal's complex processing is thus fully linear, preventing it from scaling to large datasets (See Table \ref{table:conv_compare_all}).

We introduce a general class of equivariant non-linearities which act on the relative phase information between features (Fig. \ref{fig:enql}). Given a complex-valued input feature map $\mathbf{f} \in \mathbb{C}^{C \times H \times W}$ with $C$ channels and $H \times W$ pixels, and given any non-linear activation function $\mathcal{N}: \mathbb{C} \to \mathbb{C}$, we compute an equivariant version of $\mathcal{N}$ (denoted $\mathcal{E}\{\mathcal{N}\}$) as:
\begin{align}
\mathbf{f}_{out} = \mathcal{E}\{\mathcal{N}\}(\mathbf{f}) = \mathbf{\hat m} \odot\mathcal{N}(\mathbf{f} \odot \mathbf{\hat m^*}) 
\end{align}
where $\odot$ denotes element-wise multiplication, $\hat{\mathbf{m}}^*$ denotes the complex conjugate of $\hat{\mathbf{m}}$, and $\mathbf{\hat m}$ is the normalized per-pixel mean feature:
\begin{align}
\mathbf{m}(x,y) = \frac{1}{C} \sum_{c=1}^C \mathbf{f}(c,x,y); \ \mathbf{\hat m}(x,y) = \frac{\mathbf{m}(x,y)}{|\mathbf{m}(x,y)|}
\end{align}
The normalized mean $\mathbf{\hat m}$ is equivariant to input phase and invariant to input magnitude. As a result, the product $\mathbf{f} \odot \mathbf{\hat m^*}(x,y)$ is invariant to phase and equivariant to magnitude. If $\mathcal{N}$ is equivariant to magnitude (e.g., \crelu), the overall layer $\mathcal{E}\{\mathcal{N}\}$ is equivariant to both phase and magnitude.
\figeqnl{t}

\subsection{Equivariant Pooling}
In real-valued networks, max-pooling selects the largest activations from a set of neighboring activations. However, for complex numbers, this scheme selects for phase $ 0 \leq \phi \leq \frac{\pi}{2} $ over other phases even though they may encode similar salience. Additionally, this process destroys complex-scale equivariance. To remove this dependence on phase, we select the pixels with the highest magnitude. The result is equivariant to both magnitude and phase.

\subsection{Phase-Equivariant Batch Normalization}
We follow Deng et. al \cite{vn2021}, computing Batch Normalization \cite{batchnorm} only on the magnitude of each complex-valued feature, thus preserving the phase information. Given an input feature map $\mathbf{f} \in \mathbb{C}^{C \times H \times W}$, we compute:
\begin{equation}
    \mathbf{f}_{BN} = BN(|\mathbf{f}|) \odot \frac{\mathbf{f}}{|\mathbf{f}|}
\end{equation}
where $BN$ refers to real-valued BatchNorm and $\odot$ is elementwise multiplication. This layer is equivariant to phase and invariant to input magnitude. 

\subsection{Complex-Valued Invariant Layers}

In order to produce invariant complex-valued features, we introduce the \textbf{Division Layer} and the \textbf{Conjugate Multiplication Layer}. Given two complex-valued features $\mathbf{z_1},\mathbf{z_2} \in \mathbf{C}^{H \times W}$, we define:
\begin{align}
    \text{Div}({\bf z_1,z_2}) =&
\ \frac{|\mathbf{z_1}|}{|\mathbf{z_2}| + \epsilon} \exp\{i(\ang{\mathbf{z_1}}-\ang{\mathbf{z_2}})\} \\
     \text{Conj}({\bf z_1,z_2}) =& \ \mathbf{z_1}\mathbf{z_2^*}
\end{align}
In practice, the denominator for division can be small, so we offset the magnitude of the denominator by $\epsilon = 10^{-7}$.

While the division layer induces invariance to all complex-valued scaling, the conjugate layer only induces invariance to phase. This layer also captures some second-degree interactions similar to a bilinear layer \cite{gao2016compact}. In contrast to our layers which capture relative phase and magnitude offsets of input features, SurReal's Distance Layer achieves invariance by extracting real-valued \textit{distances} between features, discarding rich complex-valued information in the process.

\figDivCinv{t}

\subsection{Generalized Tangent ReLU}
In practice, TReLU slows down convergence compared to \crelu. We remedy this through three modifications: \textbf{a)} a learned complex-valued scaling factor for each input channel, enabling the layer to adapt to input magnitude and phase, \textbf{b)} hyperparameter $r$ to control the magnitude threshold. Notably, $r=0$ produces a phase-only version of TangentReLU, which is equivariant to input magnitude, and \textbf{c)} learned scaling constant for the output phase of each channel, allowing the non-linearity to adapt the output phase distribution. Our proposed method generalizes TReLU both as a transformation and as a thresholding function. It is defined as:
\begin{align*}
    \operatorname{GTRelu}(\mathbf{x}; r,c,\omega) = \max(r, |c \cdot \mathbf{x}|) exp\{i \omega \ang{(c \cdot \mathbf{x}})_+\}
\end{align*} 
where $x \in \mathbb{C}$ is a scalar input, $r \in \mathbb{C}$ is the threshold parameter, $c \in \mathbb{C}$ and $\omega \in \mathbb{R}$ are learned scaling factors,  and $x_+ = ReLU(x) = max(x,0)$ (Figure \ref{fig:gtrelu}).

\figgtrelu{t}

\subsection{Complex Features $\to$ Real-Valued Outputs }

In order to produce real-valued outputs from complex-valued features, we propose using feature distances, which are commonly used for prototype-based classification. Given a complex-valued feature vector $\mathbf{f} \in \mathbf{C}^D_{emb}$ with embedding dimension $D_{emb}$, we compute the distance of $\mathbf{f}$ to every class prototype vector $\mathbf{p_i} \in \mathbf{C}^D_{emb}$. The input is then classified as belonging to the class of the nearest prototype $\mathbf{p_i}$. Formally, we define the logits returned by the network as the negative distance to the prototype scaled by a learned distance scaling factor $\alpha \in \mathbf{R}^+$. Formally, the i-th logit $L_i \in \mathbf{R}$ is: \begin{equation}
\label{eq:softmax}\displaystyle
L_i = - \alpha \cdot d\left(\mathbf{f}, \mathbf {p_i}\right)\end{equation}
Since the features are complex-valued, a suitable metric is the manifold distance (equivalent to \cite{chakraborty2019surreal}):
\begin{equation}
\label{eq:mdist}\displaystyle
d\left({z}_1, {z}_2\right) 
\!=\! \sqrt{\big(\ln{|z_1|}-\ln{|z_2|}\big)^2+
\arcdist(\ang{ z_1}, \ang{ z_2})^2}
\end{equation}
where $z_1, z_2 \in \mathbb{C}$. It amplifies the effect of phase differences which would otherwise be suppressed by large variations in magnitude. Alternatively, a simple metric is the Euclidean distance. 
In practice, we use BatchNorm on the input features before computing distances to accelerate convergence.

\textbf{Invariant Distance Layer}: Similar to the Distance Layer \cite{chakraborty2019surreal}, this layer can be made complex-scale invariant by multiplying the prototypes with an equivariant feature map:
\begin{equation}
\displaystyle
L_i = - \alpha \cdot d\left(\mathbf{f}, \mathbf {p_i} \odot \mathbf{m}\right)
\end{equation}
where $\mathbf{m}$ is the mean activation averaged over channels.

\subsection{Composing Equivariant and Invariant Layers}
We introduce two patterns of model composition based on our proposed layers. \textbf{Type I} models use a complex-valued \textbf{invariant} layer in the early stages to achieve complex-scale invariance, and \textbf{Type E} models use \textbf{equivariant} layers throughout the model, retaining more information.

\textbf{Type I}: These models consist of a complex-valued invariant layer (Division/Conjugate) in the early stages, producing complex-scale invariant features which can be used by later stages without any architectural restrictions.

\textbf{Type E}: These models use equivariant layers throughout the network, relying on Equivariant convolutions, non-linear activations, and pooling layers, followed by a invariant distance layer to obtain invariant predictions. This model preserves the phase information through equivariant layers and thus can typically achieve higher accuracy. However, this class of models is less flexible than Type I.

\figmodels{t}

\section{Complex-Valued Color Encodings}

Spectral representations like the Fourier Transform are invaluable for analyzing real-valued signals. However, their usefulness in convolutional neural networks is undermined by the fact that CNNs are designed to tackle spatially homogeneous and translation-invariant domains, while Fourier data is neither. We propose two \textbf{complex-valued color encodings} which capture hue shift and channel correlations respectively, demonstrating the utility of complex-valued representations for real-valued data. 

Our first so-called "Sliding" encoding takes an $[R,G,B]$ image and encodes it with two complex-valued channels: 
\begin{align}
[R, G, B] \rightarrow\left[R+iG, G+iB\right]
\end{align}
The complex phase in this encoding corresponds to the ratio of R, G, B values, so the phase in this encoding captures the correlation between the adjacent color channels.

Our second proposed encoding uses \lab , a perceptually uniform color representation with luminance represented by the $L$ channel and chromaticity by the $a$ and $b$ channels. \cite{zhang2016colorful} uses this color space for image colorization. We use it to represent color as a two-channel, complex-valued representation, with the first channel containing the luminance ($L^*$ channel), and the second channel containing chromaticity ($a^*$ and $b^*$ channels) as $a^* +i\, b^*$ (Figure \ref{fig:labex}):
\begin{equation}
[R, G, B] \rightarrow\left[L^{*}, a^{*}+i b^{*}\right]
\end{equation}
{\bf Color distortions as co-domain transformations}:
Color distortion can be approximated with complex-valued scaling of our LAB color representation (Figure \ref{fig:colorvar}).  A complex-scale invariant network is thus automatically robust to color distortions without the need for data augmentation.

%% file: 4experiments.tex
\section{Experiments}

We conduct three kinds of experiments: \textbf{Accuracy}: \textbf{1)} Classification of naturally complex-valued images, \textbf{2)} real-valued images with real and complex representations; \textbf{Robustness} against complex scaling and color distortion; \textbf{Generalization:} {\bf 1)} Bias-variance analysis, \textbf{2)} generalization on smaller training sets, {\bf 3)} Feature redundancy analysis.

\subsection{Complex-Valued Dataset: MSTAR}
MSTAR contains 15,716 complex-valued synthetic aperture radar (SAR) images divided into 11 classes \cite{keydel1996mstar}. Each image has one channel and size $128\times128$. We discard the last "clutter" class and follow \cite{wang2019successive}, training on the depression angle $17^\circ$ and testing on $15^\circ$. We train each model on varying proportions of the dataset to evaluate the accuracy and generalization capabilities of each model.

{\bf SurReal}: We replicate the architecture described in Table 1 of \cite{chakraborty2019surreal}. Since the paper does not mention the learning rate, we use the same learning rate and batch size as our model. {\bf DCN}: We use author-provided code\footnote{https://github.com/ChihebTrabelsi/deep\_complex\_networks}, creating a complex ResNet with \crelu{} and 10 blocks per stage. By default, this model accepts $32\times32$ images, so we append  $2\times [\textit{ComplexConv},\textit{ComplexBatchNorm}]$ with stride $2$ to downsample the input. The model is trained for 200 epochs using SGD with batch size $64$ and the learning rate schedule in \cite{trabelsi2017deep}. We select the epoch with the best validation accuracy. \textbf{Real-valued baseline}: We use a 3-stage ResNet with 3 layers per residual block and convert the complex input into two real-valued channels.

{\bf CDS}: We use a \textit{Type I} model based on SurReal  \cite{chakraborty2019surreal}. We extract equivariant features using an initial equivariant block containing \textit{EConv}, \textit{Eq. GTReLU}, \textit{Eq. MaxPool} layers,  and then obtain complex-scale invariant features by using a \textit{Division Layer}. These features are then fed to a real-valued ResNet.  For more details about every model, please refer to supplementary materials.

{\bf Training:}  We optimize both SurReal and CDS models using the AdamW optimizer \cite{kingma2017adam, loshchilov2018decoupled} with learning rate $10^{-3}$, momentum $(0.9,0.99)$, weight decay $0.1$, and batch size $256$ for $2.5 \times 10^5$ iterations . We validate every $1000$ steps, picking the model with the best validation accuracy. 

\figGeneralization{t}

\begin{table*}[t]
\begin{center}
\small
\begin{tabular}{c|r|lll|lll|lll}
\toprule
Method                & \multicolumn{1}{c|}{\# Param} & \multicolumn{3}{c|}{CIFAR10}                                                     & \multicolumn{3}{c|}{CIFAR100}                                                    & \multicolumn{3}{c}{SVHN}                                                        \\
\midrule
\multicolumn{1}{l|}{} & \multicolumn{1}{l|}{}              & \multicolumn{1}{c}{RGB} & \multicolumn{1}{c}{LAB} & \multicolumn{1}{c|}{Sliding} & \multicolumn{1}{c}{RGB} & \multicolumn{1}{c}{LAB} & \multicolumn{1}{c|}{Sliding} & \multicolumn{1}{c}{RGB} & \multicolumn{1}{c}{LAB} & \multicolumn{1}{c}{Sliding} \\ 
\midrule
DCN \cite{trabelsi2017deep}                  & 66,858                              & 65.17                   & 58.64                   & 63.83                        & 32.52                   & 27.36                   & 28.87                        & 85.26                   & 84.43                   & 87.44                       \\
SurReal \cite{chakraborty2019surreal}               & 35,274                              & 50.68                   & 53.02                   & 54.61                        & 23.57                   & 25.97                   & 26.66                        & 80.51                   & 53.48                   & 80.79                       \\
Real-valued CNN       & 34,282                              & 64.43                   & 63                      & 63.43                        & 31.93                   & 31.72                   & 31.93                        & 87.47                   & 84.93                   & 87.37                       \\
\midrule
Ours (Type-I) (Original \footnotemark[2])      & 24,241                              & {69.23}          & 67.17                   & 68.7                         & 36.92                   & 37.81                   & 38.51                        & {89.39}          & {88.86}          & \textbf{{90.25}}              \\
Ours (Type-I) (Corrected PGM)  & 24,241  & \textbf{70.29}	& 68.02	& \textbf{ 69.63}	& 40.79	& 38.59	& 39.4	& 89.21	& 88.98	& 89.61 \\
Ours (Type-E) (Original)    & 25,745  & 68.48                   & {67.58}          & {69.19}               & \textbf{{41.83}}          & {39.55}          & \textbf{{42.08}}   & 77.19                   & 74.21                   & 88.39     \\
Ours (Type-E) (Corrected PGM)   & 25,745   & 68.14 &	\textbf{69.38}	& 69.48 & 	41.44	& \textbf{40.53}	& 40.57 & \textbf{89.67}	& \textbf{89.5}	& 90.1 \\

\bottomrule
\end{tabular}
\end{center}
\vspace*{-5mm}
\caption{Our models outperform the baselines' CIFARnet versions on real-valued datasets. \textit{Type-I} model performs better on easier datasets like SVHN, and \textit{Type-E} performs better on difficult datasets like CIFAR100. In contrast, SurReal does not scale to large datasets.}
\label{table:conv_compare_all}
\end{table*}

\subsection{Real-valued Datasets: CIFAR10/100, SVHN}


\textbf{Datasets}: CIFAR10 \cite{krizhevsky2009learning} (and CIFAR100) consists of 10 (100) classes containing 6000 (600) images each. Both CIFAR10 and CIFAR100 are partitioned into 50000 training images and 10000 test images. SVHN \cite{netzer2011reading} consists of house number images from Google Street View, divided into 10 classes with $73,257$ training digits and $26,032$ testing digits. 

{\bf Models:}  To ensure equal footing for each model, all networks in this experiment are based off CIFARNet, i.e., 3 Convolution Layers (stride 2) and 2 fully connected layers. We also replace average pooling with a depthwise-separable convolution as a learnable pooling layer. All models are optimized with AdamW \cite{kingma2017adam, loshchilov2018decoupled} using momentum $(0.99,0.999)$, for $5\times10^4$ steps with batch size $256$, learning rate $10^{-3}$, weight decay $0.1$, and validated every $1000$ iterations. \textbf{DCN}: We use \textit{ComplexConv} for convolutions and \crelu ~as the non-linearity. We do not use Residual Blocks or Complex BatchNorm from \cite{trabelsi2017deep} to ensure fairness. \textbf{SurReal}: We use wFM for convolutions and use \textit{Distance Transform} after Layer 3 to extract invariant real-valued features. \textbf{Real-Valued CNN}: We use the CIFARNet architecture, converting each complex input channel into two real-valued channels. \textbf{CDS}: We evaluate two models: \textbf{Type I}: We use \textit{EConv} for convolutions and \textit{GTReLU} ($r=0$) for non-linearity.  We use a Division layer after the first Econv to achieve invariance. The final fully-connected layer is replaced with \textit{Prototype Distance} layer to predict class logits (Figure \ref{fig:models}). \textbf{Type E}: We use \textit{Econv} for convolutions and \textit{Equivariant GTReLU} for non-linearity. The final FC layer is replaced with \textit{Invariant Prototype Distance} layer to predict logits (Figure \ref{fig:models}), and the prototype distance inputs are normalized with \textit{Equivariant BatchNorm} to preserve equivariance.

\footnotetext[2]{PGM refers to ``phase gradient mask". Original code had a gradient computation bug in the phase thresholding in GTReLU. See \cref{sec:pgm}.}

\textbf{CDS-Large}: We train a $1.7M$ parameter \textit{Type I} model on CIFAR 10 with LAB encoding, and compare it against equivalently sized DCN (WS with \crelu ~from \cite{trabelsi2017deep}). CDS-Large is based on the simplified 4-stage ResNet provided by Page et al. \cite{pageresnet} for DAWNBench \cite{dawnbench}. We use the conjugate layer after the first Econv to get complex-scale invariant features and feed them to the Complex ResNet. Like DCN, we optimize the model using SGD with horizontal flipping and random cropping augmentations with a varying learning rate schedule (see supplementary material for more details).

\subsection{Model Performance Analysis}
{\bf Accuracy and scalability}:
Our approach achieves complex-scale invariance of manifold-based methods while retaining high accuracy and scalability. On \textbf{MSTAR}, our model beats the baselines across a diverse range of splits with less than half the parameters used by SurReal (Table \ref{table:mstar_result}). On the smallest training split ($5\%$ training data), our model shows a gain of $19.7\%$ against DCN and real-valued CNN and $8.4\%$ against SurReal. On the largest split ($100\%$ training data), our model beats real-valued CNN by $29.2\%$, DCN by $7\%$, and SurReal by $1.2\%$, demonstrating our advantage across an extensive range of dataset sizes.

\begin{table}[t]
\begin{center}
\small
\scalebox{0.87}{

\begin{tabular}{c|r|ccccc}
\toprule
Model              & \# Params & 5\%  & 10\% & 50\% & 90\% & 100\%         \\ \midrule
Real        & 33,050    & 47.4 & 46.6 & 60.6 & 73   & 66.9          \\ \midrule
SurReal \cite{chakraborty2019surreal}            & 63,690    & 61.1 & 68   & 90.3 & \textbf{95.6} & 94.9          \\ \midrule
DCN \cite{trabelsi2017deep}                & 863,587   & 49.8 &   47   & 81.9 &  89.1 & 89.1          \\ \midrule
Ours (Corrected PGM)            & 29,536    & {55.1} & {69.7} & \textbf{92.3} & 95.5 & {95.1} \\ \midrule
Ours  (Original) \footnotemark[3]          & 29,536    & \textbf{69.5} & \textbf{78.3} & {91.3} & 95.2 & \textbf{96.1} \\ 
 \bottomrule
\end{tabular}
}
\end{center}
\vspace*{-5mm}
\caption{Our method achieves the best accuracy and generalization with the fewest parameters.  We report accuracy on varying proportions of MSTAR training data. The performance gap is wider for smaller train-sets, with Real-CNN and DCN failing to generalize.}
\label{table:mstar_result}
\end{table}

\footnotetext[3]{PGM refers to “phase gradient mask". Original code had a gradient computation bug in the phase thresholding in GTReLU. Surprisingly, this made the model more robust to phase noise in the MSTAR dataset, and thus generalize better for small dataset sizes. See Sec. 6.5 in Appendix.}

On \textbf{CIFAR10, CIFAR100, and SVHN} under different encodings, our models obtain the highest accuracy across every setting (Table \ref{table:conv_compare_all}). Unlike SurReal, our model scales to these large classification datasets while retaining complex-scale invariance. For the complex-valued color encodings, which require precise processing of phase information, our model consistently beats baselines by $4\%$-$8\%$. These results highlight the advantage of our approach for precise complex-valued processing across a variety of real-valued datasets.

\figBigModelsNew{t}

{\bf Phase normalization and color jitter}:
A natural pre-processing trick to address complex-scaling invariance is to compute the average input phase $\hat \phi$ and to scale the input by $e^{-\iota \hat\phi}$ to cancel it. We test this approach by applying random complex-valued scaling with different rotation ranges and comparing DCN's accuracy with and without phase normalization against our method (Figure \ref{fig:PN}). When the input phase distribution is simple (e.g., phase set to 0), phase normalization successfully protects DCN against complex-valued scaling. However, for complicated phase distributions such as LAB encoding, this method fails. Our method succeeds in both situations, and this robustness transfers to the color jitter (as used by \cite{mean_teach}, see Figure \ref{fig:coljit}). Our model is more robust against color jitter without data augmentation. 

{\bf Bias and variance analysis}: While model accuracy across different datasets is useful, a better measure for the generalization of supervised models is the bias-variance decomposition. We follow \cite{wang2021longtailed}: given model \(f\), dataset \(D\), ground truth \(Y\) and instance \(x\), \cite{wang2021longtailed} defines the bias-variance decomposition of the prediction error (per instance) as: 
\begin{align}
    \operatorname{Error}(x ; f) &= E\left[(f(x ; D)-Y)^{2}\right]\\
    &= \operatorname{Bias}(x ; f)+\operatorname{Var}(x ; f)+\text {R}(x)
\end{align}
where bias measures the accuracy of the predictions with respect to the ground-truth, variance measures the stability of the predictions, and $R$ denotes the irreducible error. Using the 0-1 loss \(\mathcal{L}_{0-1}\), \cite{wang2021longtailed} calculates the bias and variance terms (per instance per model) for the classification task as such:
\begin{align}
\operatorname{Bias}(x ; h) &=\mathcal{L}_{0-1}\left(y_{m} ; t\right)\\
\operatorname{Var}(x ; h) &= \frac{1}{n} \sum_{k=1}^{n}\mathcal{L}_{0-1}\left(y^{(k)} ; y_{m}\right)
\end{align}
where $y_m$ is the \textit{mode} or the \textit{main prediction}. We compute this metric for each instance, averaging bias and variance over classes. Compared on CIFAR10 with LAB encoding, our model achieves the lowest bias among all classes and the lowest variance among 9 out of 10 classes (Figure \ref{fig:generalization}).

\textbf{Generalization from less training data}: \cite{kaplan2020scaling} derives empirical trends for scaling of language models under different conditions, including the overfitting regime where the training set is small compared to parameters. We produce similar trend curves for the MSTAR test results by fitting linear regression curves to log accuracy and dataset size reported in Table \ref{table:mstar_result}. We plot the results in Figure \ref{fig:generalization}. The extrapolated least-squares linear fit suggests our model might continue to generalize better on yet smaller datasets. 

\textbf{Feature redundancy comparison}: \cite{wang2020tied} shows that common CNN architectures learn highly correlated filters. This increases model size and reduces the ability to capture diversity. We follow \cite{wang2020tied}, measuring correlations between guided backpropagation maps of different filters in layer $2$ for each model on CIFAR10 with the LAB encoding. We find that our model displays the highest filter diversity. This observation is consistent with higher test accuracy, lower bias and variance, and leaner models from previous experiments.

\textbf{Scaling to large models}: Small models are essential for applications such as edge computing, thus motivating leaner models (Table \ref{table:conv_compare_all} and \ref{table:mstar_result}). However, the best models on large real-valued datasets, like ViT-H/14 \cite{dosovitskiy2021an} with $99.5\%$ test accuracy on CIFAR10, have millions of parameters. We test the scalability of our approach by comparing CDS-Large with a DCN model of equivalent size on CIFAR10 with LAB encoding (Table \ref{fig:cifar10_result}).  While we focus on leaner complex-scale invariant models, our method beats DCN while additionally achieving complex-scale invariance even for large models. This observation is consistent with our results for small models (Table \ref{table:conv_compare_all}), showing the effectiveness of our method for diverse model sizes.

\textbf{Summary}: We analyze complex-scaling as a co-domain transformation and derive equivariant/invariant 
versions of commonly used layers. We also present novel complex encodings. Our approach combines complex-valued algebra with complex-scaling geometry, resulting in leaner and more robust models with better accuracy and generalization.

\textbf{Acknowledgements}: The authors gratefully acknowledge DARPA, AFRL, and NGA for funding various applications of this research. We also thank Matthew Tancik, Ren Ng, Connelly Barnes, and Claudia Tischler for their thoughtful comments. 

%% file: 6supp.tex
\section{Supplementary Material}
\maketitle

\textbf{Table of contents:}
\begin{enumerate}[itemsep=-0.5ex,partopsep=1ex,parsep=1ex]
\item Extensive bias-variance evaluation, including plots for both LAB and RGB encodings.
\item Limitations of wFM as a convolutional layer in CNNs.
\item Ablation tests
\item Architecture details for models used in our experiments.
\end{enumerate}

\subsection{Bias-Variance Evaluation}
In this section, we run extensive evaluations for bias and variance on each encoding for the CIFAR10 dataset with CIFARNet models (Figure \ref{fig:clari}). In each case, our model (Type E) achieves the lowest bias for each class. We also achieve the lowest variance for 8 (out of 10) classes for RGB, 9 classes for LAB, and all classes for the Sliding encoding. Our overall bias and variance are the lowest of all models, indicating higher generalization ability. 
\figClarification{t}

\subsection{Limitations of wFM}
We demonstrate the theoretical and empirical limitations of wFM. From a theoretical perspective, we show that wFM using the Manifold Distance Metric of SurReal processes magnitude and phase separately and is thus unable to process the joint distribution. Our experiments show that in practice, wFM results in a significant loss in accuracy compared to real-valued and complex-valued convolutional filters.

\textbf{Decomposability}: We discuss the weighted-Fr\'echet Mean for complex-valued neural networks and show that the magnitude and phase computations are decoupled. wFM is defined as the minimum of weighted distances to a given set of points. In specific cases (like the Euclidean distance metric), closed-form solutions (like the euclidean weighted mean) exist, but there is no general closed-form solution.

Given $\left\{\mathbf{z}_i\right\}_{i=1}^K \subset \mathbf{C}$ and $\left\{w_i\right\}_{i=1}^K \subset (0,1]$ with $\sum_i w_i=1$ and $b \in \mathbf{R}$, we aspire to compute:
\begin{align}
\label{theory:eq1}
 \arg\min_{\mathbf{m}\in \mathbf{C}} \sum_{i=1}^K w_id^2\left(\mathbf{z}_i, \mathbf{m}\right)
\end{align}
Using the Manifold distance metric, the expression reduces to: 
\begin{align}
\label{theory:eq2}
 \arg\min_{\mathbf{m}\in \mathbf{C}} \sum_{i=1}^K w_i \left( (\ln|\mathbf{z}_i| - \ln |\mathbf{m}|)^2 + \arcdist^2(\mathbf{z_i},\mathbf{m})\right)
\end{align}
Note that the objective is a sum of $\log|\mathbf{z}_i| - \log |\mathbf{m}|)^2$ and $\arcdist^2(\mathbf{z_i},\mathbf{m})$, where the first objective depends only on the magnitude of $\mathbf{m}$, and the second objective depends only on the phase of $\mathbf{m}$. Thus, the magnitude and phase can be solved independently of each other. The first can be further simplified:
\begin{align}
r^* = \arg\min_{r \in \mathbf{R}} \sum_{i=1}^K w_i (\log|\mathbf{z}_i| - r)^2 \\
\implies r^* = \sum_{i=1}^K w_i . \log|\mathbf{z}_i| \\
\implies \log |m^*| = \sum_{i=1}^K w_i . \log|\mathbf{z}_i|
\end{align}
Thus, wFM simply computes a weighted sum of log magnitudes and solves a different minimization problem to find the phase. This restricts the representational power of a wFM layer compared to that of a convolution.

\textbf{wFM experiments}: We compare wFM against real-valued and complex-valued convolutions on CIFAR10 using a CIFARnet architecture. We find that the lower representational capacity of wFM leads to significant reductions in accuracy (Table \ref{table:WFM}).

\tabWFM{t}

\subsection{Ablation tests}
In this section, we run ablation tests for our \textit{Type I} model on CIFAR10 under the LAB encoding to measure the impact of layer choices on final performance. Specifically, we benchmark the Complex Invariant Layer, the Invariant Distance Layer, and different thresholds for the Generalized Tangent ReLU layer (Table \ref{table:ablations}). We find that the Division layer beats the Conjugate layer in terms of accuracy and that Manifold Distance achieves higher accuracy than our baseline of Euclidean Distance. We also note that among GTReLU thresholds, $r=0.1$ performs the best but is not equivariant to magnitude. We also note that higher thresholds like $r=1$ result in lower accuracy.

\tabAblations{t}

\subsection{Architecture Details}
In this section, we discuss details of the architectures used in our experiments. 

\textbf{CIFARnet architectures}: For CIFARnet architectures, please refer to tables \ref{tab:SurRealCifarNet}-\ref{tab:RealCifarNet}. Please note that our replication of wFM \cite{chakraborty2019surreal} uses the $(\log mag, sin \theta, cos \theta)$ encoding for the manifold values, and uses the weighted average formulation.

\textbf{MSTAR architectures}: For DCN, please refer to \cite{trabelsi2017deep}, and for the downsampling block, see Table \ref{tab:DCNmstar}. Our SurReal replication is based on Table I in \cite{chakraborty2019surreal}, and our model is based on the SurReal architecture (see Table \ref{tab:ourMSTAR}). We use the same real-valued ResNet as the real-valued baseline (see Table \ref{tab:realmstar}). In order to pass the complex-valued features into the real-valued ResNet, we convert complex features to real-valued using the $(\log mag, sin \theta, cos \theta)$ encoding, treating each as a separate real-valued channel (resulting in 15 real-valued channels from 5 complex-valued channels).

\textbf{CDS-Large}: For the model architecture, please see Table \ref{tab:ourBIG}. We train this model with SGD, using momentum $0.9$, weight decay constant $5 \times 10^{-4}$, using a piece-wise linear learning rate schedule starting at $0.01$, increasing to $0.2$ by epoch $10$, then decreasing to $0.01$ by epoch $100$, $0.001$ by $120$, $0.0001$ by $150$, and staying constant until $200$. To ensure fair comparison, use horizontal flips and random cropping augmentation as used in \cite{trabelsi2017deep}. All models are implemented in PyTorch \cite{NEURIPS2019_9015}.

\subsection{Errata: Phase Gradient Mask} \label{sec:pgm} 

\begin{table*}[t]
    \centering
    \begin{tabular}{c|cccc}
        \toprule
        Code & Forward Pass & Magnitude Gradient & Phase Thresholding gradient (value) & Phase Thresholding gradient (mask) \\
        \midrule
         Original & \checkmark & \checkmark & \checkmark & $\operatorname{Mask} = 1-(0^\circ\leq \operatorname{Phase} \leq 180^\circ)$\\
         Correct & \checkmark &  \checkmark & \checkmark & $\operatorname{Mask} = 0^\circ\leq \operatorname{Phase} \leq 180^\circ$\\
         \bottomrule
    \end{tabular}
    \caption{Gradient computation bug in GTReLU computed the wrong mask for phase thresholding gradients. The values were correct, but the mask was flipped. As a result, the phase gradients for phase thresholding stage were enabled for points in the lower half of the complex plane rather than the upper half of the complex plane.}
    \label{tab:errata}
\end{table*}

This code uses a version of the Tangent ReLU non-linearity originally proposed by SurReal. However, the implementation we used had flipped the phase gradient mask for phase thresholding. Specifically, the forward pass was correct, and the magnitude gradients were correct, but the backward pass for the phase thresholding allowed phase gradients through for $\theta < 0$ angles rather than $\theta > 0$ angles. The forward pass of phase thresholding in both cases is still $x_{+}$, but the backward pass mask was previously $\mathbf{1}\{ x < 0 \}$ rather than $\mathbf{1}\{ x \geq 0 \}$. See \cref{tab:errata} for details.

\textbf{Result:} Surprisingly, this modification increases the accuracy for MSTAR (especially for smaller dataset sizes). The true gradient mask, in contrast, has better convergence and stability (and higher accuracy for CIFAR experiments). This difference highlights the strength of each type of phase mask. For completeness, we have included results for the original (``flipped" phase gradient mask) and the true gradient mask.

\textbf{Impact on phase manipulation}: GTReLU has multiple mechanisms for controlling the phase of each input. Since the learned scaling factor (i.e., the stage before thresholding) and phase-scaling (i.e., the stage after thresholding) control the phase in a learnable manner, the model can still manipulate the input phases regardless of the phase gradient mask. This explains the small effect of the bug in most settings.

PGM mainly decides which input phases change individually (i.e., by backpropagating individual phase gradients) and which phases only undergo global phase shift/scaling (i.e., through phase-scaling and learned scaling factor). The correct PGM allows $0 \leq \theta \leq 180$ phases to change individually while the rest undergo global shift/scaling. The flipped phase gradient mask does the opposite, allowing phases clipped to 0 to get individual gradients. 

While the exact orientation in the complex plane (e.g., $0 \leq \theta \leq 180$) is meaningless due to learned scaling and $\operatorname{CConv}$ layers, clipped input phases are \textit{statistically} different from non-clipped inputs since clipping concentrates the values on the positive-real line. If the original phase was noisy (e.g., in MSTAR), the flipped PGM may eliminate the impact of noisy phase inputs on the backpropagated gradient. Our current understanding is that the increased MSTAR accuracy for the "flipped phase gradient mask'' may come from this increased phase robustness. In contrast, the individual phase control offered by the correct PGM may allow for better convergence and accuracy for CIFAR models.

\tabSurRealCifarNet{t}
\tabDCNCifarNet{t}
\tabourECifarNet{t}
\tabourICifarNet{t}
\tabRealCifarNet{t}
\tabourBIG{t}
\tabDCNmstar{t}
\tabrealmstar{t}
\tabourMSTAR{t}